%% file: main.tex
\documentclass[sigconf]{acmart}

\AtBeginDocument{%
  \providecommand\BibTeX{{%
    \normalfont B\kern-0.5em{\scshape i\kern-0.25em b}\kern-0.8em\TeX}}}

\copyrightyear{2026}
\acmYear{2026}
\setcopyright{cc}
\setcctype[4.0]{by}
\acmConference[CAIS '26]{ACM Conference on AI and Agentic Systems}{May 26--29, 2026}{San Jose, CA, USA}
\acmBooktitle{ACM Conference on AI and Agentic Systems (CAIS '26), May 26--29, 2026, San Jose, CA, USA}
\acmDOI{10.1145/3786335.3813154}
\acmISBN{979-8-4007-2415-2/2026/05}

\graphicspath{{./images/}} 

\usepackage{subcaption}
\usepackage{makecell}
\usepackage{cleveref}
\usepackage{bbold}
\usepackage{enumitem}
\usepackage{multirow}
\usepackage{xspace}
\usepackage{caption}
\usepackage{framed}
\newcommand{\algstyle}[1]{\texttt{#1}}
\newcommand{\ZEROSHOT}{\algstyle{zero\_shot}}
\newcommand{\EPLABEL}{\algstyle{EP\_LABEL}}
\newcommand{\EPCRIT}{\algstyle{EP\_CRIT}}

\newcommand{\SEMCRIT}{\algstyle{SEM\_CRIT}}
\newcommand{\EPSEMCRIT}{\algstyle{EP+SEM\_CRIT}}

\definecolor{tcbframe}{gray}{0.4}  
\definecolor{tcbbody}{gray}{0.96}  
\definecolor{tcbtitle}{gray}{1.0}  
\NewDocumentEnvironment{promptbox}{ O{} +m }
  {%
    \def\temp{#1}\ifx\temp\empty\else\label{#1}\fi
    \begin{framed}
    \noindent{\bfseries #2}\par
    \vspace{4pt}\hrule\vspace{6pt}
    \ttfamily\small
  }
  {%
    \end{framed}
  }

\begin{document}

\title{Learning from Supervision with Semantic and Episodic Memory: \\A Reflective Approach to Agent Adaptation}

\author{Jackson Hassell}
\affiliation{%
  \institution{Megagon Labs}
  \city{Mountain View}
  \country{USA}}
\email{jackson@megagon.ai}

\author{Dan Zhang}
\affiliation{%
  \institution{Megagon Labs}
  \city{Mountain View}
  \country{USA}}
\email{dan_z@megagon.ai}

\author{Hannah Kim}
\affiliation{%
  \institution{Megagon Labs}
  \city{Mountain View}
  \country{USA}}
\email{hannah@megagon.ai}

\author{Tom Mitchell}
\affiliation{%
  \institution{Megagon Labs}
  \city{Mountain View}
  \country{USA}}
\email{tom@megagon.ai}

\author{Estevam Hruschka}
\affiliation{%
  \institution{Megagon Labs}
  \city{Mountain View}
  \country{USA}}
\email{estevam@megagon.ai}

\begin{abstract}
We investigate how agents built on pretrained large language models (LLMs) can learn target classification functions from labeled examples without parameter updates. While conventional approaches like fine-tuning are often costly, inflexible, and opaque, we propose a memory-augmented framework that leverages LLM-generated critiques grounded in labeled data. Our framework uses episodic memory to store instance-level critiques—capturing specific past experiences—and semantic memory to distill these into reusable, task-level guidance.

Across a diverse set of tasks and models, our best performing self-critique strategy (utilizing both memory types) yields an average improvement of 8.1 percentage points over the zero shot baseline, and 4.6pp over a RAG-based baseline that relies only on labels. However, improvements vary substantially across models and domains. To explain this variation, we introduce suggestibility - a novel metric capturing how receptive a model is to external reasoning provided in context. We use suggestibility to illuminate when and why memory augmentation succeeds or falls short. Beyond accuracy gains, we find pre-computed critiques  substantially reduce inference-time computation for reasoning models, cutting thinking tokens by an average of 31.95\% across all datasets by substituting for reasoning that the model would otherwise perform independently. Our findings highlight the conditions under which memory-driven, reflective learning can serve as a lightweight, interpretable, and efficient strategy for improving LLM adaptability.

\end{abstract}



\maketitle

\input{tex/intro}

\input{tex/method}
\input{tex/exp}
\input{tex/related}
\input{tex/conclusions}

\bibliographystyle{ACM-Reference-Format}
\bibliography{references, anthology}

\input{tex/appendix}

\input{tex/artifact_appendix}
\end{document}

%% file: tex/intro.tex
\section{Introduction}
\label{sec:intro}

Large language models (LLMs) have demonstrated impressive generalization capabilities across a wide range of tasks. These AI agents rely on intelligence embedded in their pretrained parameters, and increasingly, on learning from task-specific signals, whether explicit (e.g., labeled supervision) or implicit (e.g., user interactions, feedback). A key challenge is enabling agents to continuously improve their performance and generalize to unseen domains or tasks by distilling knowledge from such signals and storing them in a reusable and interpretable form.

Traditional approaches to learning from new signals often involve updating model parameters through fine-tuning~\cite{radford2018improving, howard-ruder-2018-universal} or adaptation mechanisms such as parameter-efficient methods (e.g., LoRA adapters)~\cite{pmlr-v97-houlsby19a,hu2022lora}. While effective, these approaches incur computational cost, require retraining for every new signal or task, and often lack interpretability or controllability. Furthermore, they provide limited support for never-ending learning, where an agent must continuously adapt without retraining from scratch or storing large sets of models.

An alternative paradigm is memory-augmented learning~\cite{a27da5feb471466cb024242bf91426d3,Zhong_Guo_Gao_Ye_Wang_2024, salama-etal-2025-meminsight}, where the underlying model remains frozen, and adaptation occurs through interaction with an external memory. This memory stores relevant task knowledge, examples, demonstrations, or explanations, that can be retrieved at inference time to inform the model’s decisions. Among such approaches, in-context learning (ICL)~\cite{dong-etal-2024-survey} has emerged as a simple yet powerful mechanism, where the model is conditioned on a prompt consisting of a small number of examples (few-shot learning). 
However, directly incorporating supervised signals in the LLM context often relies on only few-shot input-output examples and tends to result in shallow pattern mimicking, due to a lack of deeper abstraction or conceptual understanding.

\begin{figure*}
    \centering
    \includegraphics[width=1.0\textwidth]{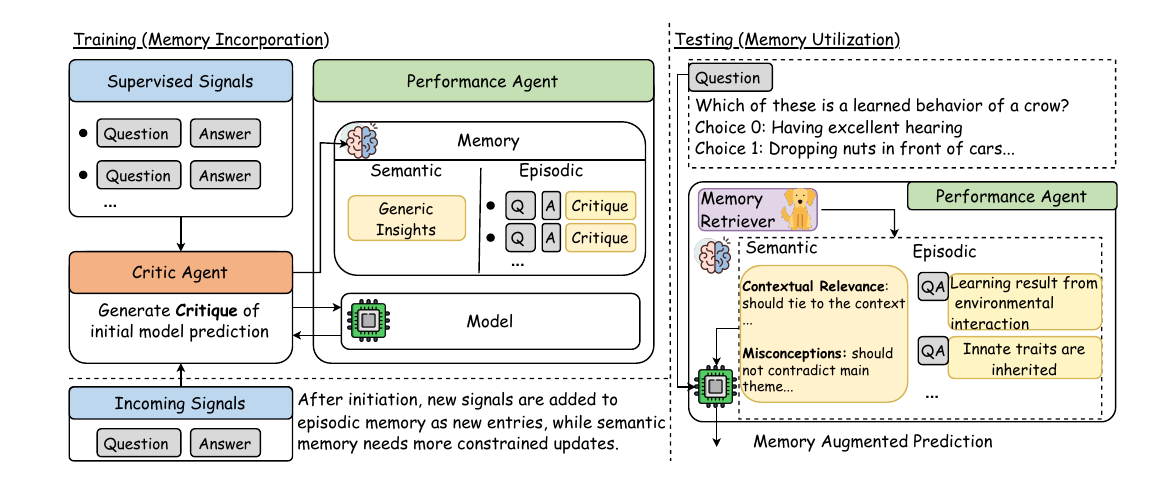}
    \caption{Agents learn from supervised signals by incorporating them into memory. At inference time, both task-level insights (semantic memory) and context-specific information (episodic memory) can support more informed decision-making. }
    \label{fig:diagram}
\end{figure*}

Recent work ~\cite{NEURIPS2023_91edff07,yao2023react,gou2024critic,NEURIPS2023_1b44b878, zhang2025spiralsymbolicllmplanning} has highlighted the capacity of LLM agents to not only perform tasks but also critique them, generating feedback and identifying patterns of errors in their own outputs. Several notable approaches leverage self-critique, but they differ fundamentally from our objectives. Self-Refine~\cite{NEURIPS2023_91edff07} relies purely on a model iteratively refining its response to a single question using parametric knowledge, and does not produce reusable memory artifacts utilized in future predictions. CRITIC~\cite{gou2024critic} extends this iterative process by grounding self-critiques in external tool outputs, but similarly lacks the generation of reusable memories derived from gold-standard data. Reflexion~\cite{NEURIPS2023_1b44b878} approximates reinforcement learning by storing self-critiques from synthetic environments, but it relies on binary correctness signals from the environment rather than pre-labeled data and restricts memory retrieval to identical tasks, limiting its ability to generalize. In contrast to these methods, our primary differentiator is grounding critiques in gold-standard labeled data to produce reusable memory artifacts. This grounds the critiques in supervised data as a way to provide novel information to the agent, avoiding the simple reinforcement of the base LLM's parametric biases.  

Inspired by human tutoring, where feedback often includes explanations of mistakes and guidance for improvement, we explore whether such reflective insights can be distilled into reusable knowledge for future tasks. Instead of merely memorizing example responses, we hypothesize that an agent that internalizes structured feedback can develop a deeper understanding of task requirements and generalize more effectively to new examples.

In this paper, we investigate how LLM agents can effectively and continuously learn from supervised signals and deeper reasoning provided by critiques, incorporating these insights into memory.

%% file: tex/method.tex
\section{Learning from Supervised Signals}
For large language model (LLM)-based agents, the availability of supervised signals, in the form of labeled datasets or continuously incoming feedback from users or the environment, presents a wide range of opportunities for building agents that can learn and improve continuously.

As illustrated in \Cref{fig:diagram}, we refer to our task-solving agent as the \textit{performance agent} (PA). The PA consists of an LLM model (the prediction model), which is queried to perform tasks, and a memory module, which it can read from and write to. Given a new task to which we want to adapt the agent, we begin with an initial labeled dataset:

$\mathcal{D}_{\text{train}}^{\text{init}} = \{(x_i, y_i)\}_{i=1}^{N}$,
where each \(x_i\) represents a task-related question or request, and \(y_i\) denotes the corresponding correct label or answer. The performance agent processes inputs \(x_i\) from a test set \(\mathcal{D}_{\text{test}}\) and produces initial predictions denoted by \(\text{PA}(x_i)\).

To enhance the capabilities of the PA, we introduce a second component: the \textit{critic agent} (CA) making use of a LLM critic model. The CA takes as input one tuple \((x_i, y_i)\) along with the PA's prediction \(\text{PA}(x_i)\), and outputs a text critique aimed at improving the PA's performance. For all experiments, the CA uses the same underlying model as the PA.

\subsection{What to Remember?}

Critique is a widely used approach for improving model performance and guiding iterative refinement by identifying errors, uncovering blind spots, and providing actionable feedback for enhancement~\cite{NEURIPS2023_1b44b878,gou2024critic,chen2024teaching, burleigh-etal-2025-beyond}.
In our setup, we employ a label-driven critique generation process, where the critic agent uses provided ground-truth answers as part of its input to generate critiques. For each question in the dataset, the performance agent first produces an initial prediction. The critic agent is then given the correct answer and asked to critique the performance agent’s output. Each critique is structured into the following fields:

\begin{itemize}[leftmargin=1.5em,topsep=0.8em]
    \item \textbf{Assertion:} A reiteration of the correct answer to the question, and a judgment regarding the correctness of the performance agent’s response.
    \item \textbf{Rationale:} An instance-specific explanation detailing why the correct answer is valid and why the performance agent’s response was correct or incorrect.
    \item \textbf{Reflection:} A broader, generalizable insight that may be applicable to similar questions in the future.
\end{itemize}

This design addresses a key challenge observed in our empirical studies: critic agents sometimes persist in their own incorrect understanding from their underlying model when generating critiques, even after being shown the correct answer. Because the critic agent draws on the model’s parametric knowledge, it can inherit pretraining biases that reinforce such biases. To mitigate this, we require the critic agent to explicitly restate the correct answer and make a clear assertion about the correctness of the initial prediction before offering a rationale or reflection. This explicit structure significantly reduces confirmation bias.

We further decompose critiques into two conceptual layers - \textit{rationale} (local) and \textit{reflection} (global) - to balance specificity and generalizability. An ideal rationale should provide a detailed explanation tailored to the specific instance, while a reflection should capture broader insights that can be applied to unseen examples in the future. Upon inspection, this structured format resulted in noticeably higher-quality critiques. See \Cref{tab:critique_parts_ablation} for an ablation study of performance with different components of the critique.

\begin{promptbox}{Example Critique}
\textbf{Question:} Does short-term treatment with proton pump inhibitors cause rebound aggravation of symptoms? \newline
\textbf{PA Response:} Yes \newline
\textbf{Critique:}
\begin{itemize}[leftmargin=.5em,topsep=0.3em]
\setlength{\itemsep}{-0.3em}
\item \textit{Assertion:} No
  \item \textit{Rationale:} Short-term treatment with proton pump inhibitors (PPIs) generally does not cause rebound aggravation of symptoms upon discontinuation. Studies have shown that any perceived increase in symptoms after stopping PPIs may be temporary...
  \item \textit{Reflection:} In the broader context, medical treatments may sometimes be misattributed with rebound phenomena, but each class of medication has its own pharmacological profiles.  In the case...
\end{itemize}
\end{promptbox}

\section{Incorporating Critiques into Memory}

Next, we investigate how learned critiques can be effectively incorporated into the performance agent’s memory. We adopt two primary forms of PA memory: \textit{semantic memory} and \textit{episodic memory}, both of which are well-established in agentic learning literature~\cite{sumers2024cognitive}. For examples of both kinds of memory, see \cref{appendix_examples}.

\subsection{Semantic Memory}

Semantic memory encodes generalizable knowledge across the entire dataset. In this work, we construct semantic memory by \textit{summarizing all of the critiques} across a dataset into a unified knowledge representation. This allows the performance agent to draw on abstract insights during inference in future tasks.
Semantic memory aims to capture insights that are broadly applicable across the entire task domain. To utilize it at inference time, we augment the performance agent’s prompt with these insights in the form of additional instructions. This strategy is referred to as \SEMCRIT{}.
Semantic memory typically takes the form of a bulleted list of task-specific advice and cautions, though its format is intentionally left vague to give the critic agent flexibility in capturing patterns from previously generated critiques.

\subsection{Episodic Memory}

While semantic memory offers concise and broadly applicable knowledge, it often fails to capture nuanced patterns or context-specific behaviors, particularly in diverse datasets. Episodic memory addresses this by enabling the agent to recall specific past instances, effectively allowing it to "revisit" similar scenarios where successes or failures occurred, along with accompanying context and critical reasoning.

The key to effective use of episodic memory lies in the retrieval of relevant examples. The agent retrieves relevant memories from similar prior cases and conditions on both the original examples and their critiques, learning to weigh and incorporate only the most pertinent critiques rather than attending to all examples equally. This strategy is referred to as (\EPCRIT{}). Following the retrieval-augmented generation (RAG) paradigm, we identify the top $K=5$ most similar data points to a test input $x_i$ using semantic embeddings. The corresponding memory entries, containing critical thinking artifacts are then used as additional demonstrations for the performance agent. See \Cref{varying_k} for detailed results and discussion on why we chose $K=5$.

\subsection{Combining Semantic and Episodic Memory}

To harness the complementary strengths of both memory types, we introduce \EPSEMCRIT{}. This hybrid strategy presents the performance agent with both high-level semantic instructions and context-specific episodic examples. By unifying generalizable semantic memory with detailed situational context, these approaches aim to support more robust and adaptive reasoning during inference. These are unified by simply concatenating the semantic memory to the end of the episodic memory.

%% file: tex/exp.tex
\section{Empirical Evaluation}

\begin{table*}[ht]
\centering
\begin{subtable}[t]{0.48\textwidth}
\centering
\scriptsize
\setlength{\tabcolsep}{2pt}
\resizebox{\linewidth}{!}{
\begin{tabular}{lccccccc}
\toprule
\makecell{Model and \\ Experiment} & \makecell{Multi-\\Cond.\\Ranking} & \makecell{NFCorpus} & \makecell{PubMed} & \makecell{Steam\\Pref} & \makecell{Book\\Pref} & \makecell{Anime\\Pref} & \makecell{Movie\\Pref} \\
\midrule
\multicolumn{8}{l}{\textbf{gpt-4o-mini-2024-07-18}} \\
\quad \ZEROSHOT{} & \underline{35.2} & 87.2 & \underline{64.0} & 58.4 & 55.6 & 56.4 & 51.6 \\
\quad \EPLABEL & 32.0 & 83.6 & 62.8 & 61.2 & 58.0 & 57.6 & 45.6 \\
\quad \EPCRIT{} & 34.8 & \underline{89.2} & 63.2 & \underline{71.2} & \underline{64.8} & \textbf{62.0} & \textbf{58.4} \\
\quad \SEMCRIT{} & 32.8 & 85.2 & 61.6 & 60.4 & 58.8 & 52.8 & 55.2 \\
\quad \EPSEMCRIT{} & \textbf{36.8} & \textbf{90.0} & \textbf{64.8} & \textbf{73.6} & \textbf{66.8} & \underline{61.6} & \underline{58.4} \\
\hline
\multicolumn{8}{l}{\textbf{llama-4-scout}} \\
\quad \ZEROSHOT{} & \textbf{38.8} & 85.6 & \textbf{72.4} & 56.4 & 56.8 & 54.0 & 49.6 \\
\quad \EPLABEL & 37.2 & 90.4 & \underline{71.2} & \textbf{66.8} & 54.4 & \textbf{60.4} & 54.8 \\
\quad \EPCRIT{} & 36.8 & \textbf{93.6} & 70.0 & 62.4 & \underline{60.8} & \underline{58.0} & \textbf{57.6} \\
\quad \SEMCRIT{} & \underline{38.8} & 83.6 & 69.2 & 63.2 & 59.2 & 56.8 & 50.4 \\
\quad \EPSEMCRIT{} & 34.8 & \underline{91.2} & 70.8 & \underline{63.6} & \textbf{61.6} & 57.6 & \underline{56.4} \\
\hline
\multicolumn{8}{l}{\textbf{qwen3-235b-a22b-instruct-2507}} \\
\quad \ZEROSHOT{} & 42.0 & 86.4 & 66.8 & 54.8 & 54.8 & 52.8 & 51.6 \\
\quad \EPLABEL & 44.0 & 94.8 & \underline{67.2} & 55.6 & 58.8 & 64.8 & 52.0 \\
\quad \EPCRIT{} & \textbf{56.8} & \textbf{95.2} & 64.8 & \textbf{68.4} & \underline{63.6} & \textbf{69.6} & 58.4 \\
\quad \SEMCRIT{} & 42.4 & 85.2 & \textbf{68.4} & 64.0 & 60.4 & 61.6 & \underline{62.8} \\
\quad \EPSEMCRIT{} & \underline{54.0} & \underline{95.2} & 64.8 & \underline{68.4} & \textbf{66.4} & \underline{65.2} & \textbf{64.0} \\
\bottomrule
\end{tabular}
}
\caption{Non-Reasoning models}
\label{tab:model_performance}
\end{subtable}
\hfill
\begin{subtable}[t]{0.48\textwidth}
\centering
\scriptsize
\setlength{\tabcolsep}{2pt}
\resizebox{\linewidth}{!}{
\begin{tabular}{lccccccc}
\toprule
\makecell{Model and \\ Experiment} & \makecell{Multi-\\Cond.\\Ranking} & \makecell{NFCorpus} & \makecell{PubMed} & \makecell{Steam\\Pref} & \makecell{Book\\Pref} & \makecell{Anime\\Pref} & \makecell{Movie\\Pref} \\
\midrule
\multicolumn{8}{l}{\textbf{gpt-5-2025-08-07}} \\
\quad \ZEROSHOT{} & 81.6 & 90.0 & 70.0 & 53.6 & 52.0 & 55.6 & 48.8 \\
\quad \EPLABEL & 95.2 & 96.4 & \underline{71.2} & 64.4 & \underline{71.2} & \underline{72.0} & \textbf{69.2} \\
\quad \EPCRIT{} & 96.8 & \textbf{97.6} & 69.2 & \textbf{69.2} & 70.0 & \textbf{73.2} & 65.6 \\
\quad \SEMCRIT{} & \underline{98.0} & 91.6 & 69.2 & 64.4 & 65.6 & 58.0 & 63.6 \\
\quad \EPSEMCRIT{} & \textbf{99.2} & \underline{96.8} & \textbf{73.6} & \underline{68.8} & \textbf{73.2} & 68.8 & \underline{67.6} \\
\hline
\multicolumn{8}{l}{\textbf{gpt-oss-20b}} \\
\quad \ZEROSHOT{} & 58.0 & 90.4 & \underline{66.8} & 52.4 & 56.4 & 52.8 & 52.0 \\
\quad \EPLABEL & 34.0 & 88.0 & 64.8 & 57.2 & 55.6 & 53.6 & 54.0 \\
\quad \EPCRIT{} & 57.6 & \underline{94.8} & 66.0 & \textbf{63.2} & 58.4 & \underline{60.8} & \textbf{64.0} \\
\quad \SEMCRIT{} & \textbf{72.0} & 89.6 & 60.4 & 57.2 & \textbf{60.0} & 60.4 & 52.4 \\
\quad \EPSEMCRIT{} & \underline{58.8} & \textbf{95.6} & \textbf{68.0} & \underline{62.8} & \underline{59.6} & \textbf{61.2} & \underline{63.6} \\
\hline
\multicolumn{8}{l}{\textbf{qwen3-vl-235b-a22b-thinking}} \\
\quad \ZEROSHOT{} & 46.2 & 87.6 & \textbf{64.8} & 58.8 & 53.6 & 55.6 & 49.2 \\
\quad \EPLABEL & 50.4 & 94.8 & 62.4 & 56.6 & \underline{62.0} & \textbf{64.8} & 61.6 \\
\quad \EPCRIT{} & 64.5 & \textbf{98.8} & \underline{62.8} & \textbf{64.4} & 57.6 & \underline{63.6} & \textbf{67.2} \\
\quad \SEMCRIT{} & \underline{65.6} & 88.4 & 60.0 & 57.2 & 58.8 & 54.4 & 62.0 \\
\quad \EPSEMCRIT{} & \textbf{72.8} & \underline{97.6} & 62.0 & \underline{62.0} & \textbf{62.0} & 63.6 & \underline{64.0} \\
\bottomrule

\end{tabular}
}
\caption{Reasoning models}
\label{tab:model_performance_reasoning}
\end{subtable}
\caption{Performance agent accuracy across datasets. We use \texttt{EP}, \texttt{SEM}, and \texttt{EP+SEM} to denote episodic, semantic, and combined memory. Results on preference datasets are averaged across all users. For each model and dataset, the highest score is \textbf{bolded} and the second-highest is \underline{underlined}.}
\end{table*}

\subsection{Datasets}

To evaluate the effectiveness of various memory-augmented learning strategies under diverse conditions, we conduct empirical studies across multiple datasets.
The tasks cover a range of settings, including fact-oriented question answering, ranking, and retrieval-based QA. The following datasets were selected based on several criteria: domain diversity, easily verifiable answers, and low enough baseline accuracy to allow room for improvement.

\paragraph{Multi-Condition Ranking~\cite{multi_condition_ranking}} Given a list of 5 items, sort them in order along 3 logical conditions. Converted into a 4-choice multiple-choice task.

\paragraph{NFCorpus~\cite{NFCorpus}} Given a medical article and two medical papers, determine which paper is cited directly by the article's bibliography.

\paragraph{PubMed~\cite{pubmed}} Determine if a highly technical medical statement is true or false, across many different medical domains. \newline

To complement these more knowledge-intensive tasks above, we also evaluated our strategies on four personal preference datasets, where models would have no prior knowledge or biases to rely on. In these tasks, the goal is to predict whether a given item belongs to a user's history. Even if the model had encountered these datasets during training, it would be unlikely to memorize preferences associated with individual user IDs, making these tasks a controlled testbed for assessing whether self-critique strategies can improve performance purely through learning from experience rather than simply leveraging preexisting knowledge.

The preference datasets were collected for four media types: \textbf{Steam Pref~\cite{SteamPref}} for video games, \textbf{Book Pref~\cite{book_preference}}, \textbf{Anime Pref~\cite{AnimePref}}, and \textbf{Movie Pref~\cite{MoviePref}}. For these datasets, we randomly selected ten users per dataset. For each user, 50 items were sampled from their history and paired with an item of equivalent popularity that was not present in their history, to prevent the model from simply guessing the more popular item for each pair. These were then formulated into 50 questions, split equally into train and test sets, of the format "Is user $user\_id$ more likely to prefer $item1$ or $item2$?"

For all other datasets, 500 questions were randomly sampled and evenly split into training and testing sets. Additional dataset-specific preprocessing details are provided in \Cref{app:data}.

We separate these datasets into two distinct groups: fact-oriented datasets, which includes Multi-Condition Ranking, NFCorpus, and PubMed, and preference-based datasets, which include the other four.

\subsection{Experimental Setup}

We compared different learning strategies against two baseline setups: \ZEROSHOT{} and \EPLABEL{}. The \ZEROSHOT{} baseline reflects the agent's performance without memory or demonstrations. The \EPLABEL{} baseline is a retrieval-based few-shot strategy that includes $K=5$ example question–answer pairs $(x_i, y_i)$ retrieved from the training set using embedding similarity, consistent with all other EP strategies, where $x_i$ is the input and $y_i$ the corresponding answer. \EPLABEL{} is a strong baseline—combining RAG and few-shot demonstration—as it has full access to the same supervised signals. While \EPLABEL{} can be categorized as a RAG solution, we adopt the EP naming convention to highlight both its similarities to and differences from other strategies.

We experiment with six different models. In \Cref{tab:model_performance}, we examine the performance of non-reasoning models, and in \Cref{tab:model_performance_reasoning} we examine the performance of specialized reasoning models.
These models span a range of sizes and include both dense and mixture-of-experts (MoE) architectures. Hyperparameters used were \texttt{token\_limit=8192}, \texttt{reasoning\_effort=medium}, \texttt{temperature=0}, for each model that supported that hyperparameter. If a model ran out of tokens before giving an answer, that answer was considered incorrect.

\subsection{Main Results}
\label{sec:main_results}

\Cref{tab:model_performance} and \Cref{tab:model_performance_reasoning} compare various learning strategies across all seven datasets. We observe that critique-based memory augmentation improves performance in the majority of model-dataset combinations, though the degree of improvement varies substantially depending on model, dataset, and method. All reported improvements and gains are in absolute percentage points.

\paragraph{Critique-based methods outperform baselines across most configurations.}
Five of our six models show improvements over the best baseline on the majority of datasets. The hybrid \EPSEMCRIT{} strategy emerges as the most consistently effective approach, averaging 8.10\% improvement over \ZEROSHOT{} averaged across all models and datasets, while \EPCRIT{} saw an average gain of 7.56\% and \SEMCRIT{} saw an average gain of 3.67\%.

The \EPLABEL{} baseline reinforces existing trends in literature, very often showing substantial improvements over \ZEROSHOT{} and averaging an improvement of 3.46\%. Even without critiques, RAG few-shot in-context learning remains a potent tool for augmenting agents with memory. However, the consistent gains of critique-based methods over \EPLABEL{} demonstrate that the structured reasoning in the critiques provides valuable signal beyond raw input-output examples.

\paragraph{Episodic memory methods often outperform \SEMCRIT{}, but semantic memory still has value.} Semantic memory typically outperforms \ZEROSHOT{}, indicating that high-level task insights help the performance agent make better decisions. In addition, while \SEMCRIT{} does not consistently beat the \EPLABEL{} baseline, \EPSEMCRIT{} is often better than pure \EPCRIT{}. 
This suggests that combining the generalization of semantic memory with the specificity of episodic memory yields more robust performance than either approach alone.

\paragraph{Performance improvement is uneven across datasets.} PubMed was the dataset that saw the least improvement for all six models, and in some cases \ZEROSHOT{} did better on PubMed than \EPLABEL{}. On the other hand, the preference datasets saw some of the largest performance improvements over baselines across models. This implies that some types of tasks see more benefit with memory augmentation than others. PubMed requires the agent to know very specific facts which are often not shared across questions, making it harder for the model to pull relevant insights from the retrieved examples. When making predictions on Book Pref, on the other hand, the model can make several deductions about a user's preferences from just a few provided examples, and these deductions can be further enhanced through structured reasoning by using the critique methods.

\paragraph{Not all models benefit equally from critique-based augmentation.} Llama 4 Scout shows the most limited improvements, outperforming baselines on only 3 of 7 datasets. In contrast, GPT-oss-20B exhibits the largest  improvement of \EPSEMCRIT{} over \EPLABEL{}, with an average gain of 8.9\% across all datasets. Reasoning models generally achieve higher baseline performance than non-reasoning models of comparable size, yet they still show substantial gains from memory augmentation.

Interestingly, while GPT-5 shows relatively modest improvements with critique-based method relative to \EPLABEL{} - averaging only 1.2\% gain - it is the model with the strongest improvement compared to \ZEROSHOT{} with an average gain of 13.8\%. 
One possible explanation is that this highly capable model already extracts substantial insight from raw examples during its thinking process, partially replicating the benefits of explicit critiques—though we cannot directly verify this without access to GPT-5's internal reasoning traces. However, if this explanation holds, we would expect explicit critiques to reduce the reasoning burden for models that would otherwise derive similar insights themselves. 
We explore the relationship between critique-based methods and inference-time computation in \Cref{sec:token_analysis}.

\begin{table}[t]
\centering
\scriptsize
\setlength{\tabcolsep}{4pt}
\begin{tabular}{llrrrrr}
\toprule
Critique & Baseline & Mean \% & SD \% & 95\% CI & +/$-$/0 & $p$ (sign) \\
\midrule
\EPCRIT{} & \ZEROSHOT{} & +7.6 & 6.6 & [+5.6, +9.7] & 33/9/0 & 0.000272 \\
\SEMCRIT{} & \ZEROSHOT{} & +3.7 & 6.3 & [+1.7, +5.7] & 28/13/1 & 0.0275 \\
\EPSEMCRIT{} & \ZEROSHOT{} & +8.1 & 6.6 & [+6.0, +10.1] & 38/4/0 & 5.65e-08 \\
\EPCRIT{} & \EPLABEL & +4.1 & 5.8 & [+2.3, +6.0] & 32/10/0 & 0.000941 \\
\SEMCRIT{} & \EPLABEL & +0.2 & 8.2 & [-2.3, +2.8] & 18/22/2 & 0.636 \\
\EPSEMCRIT{} & \EPLABEL & +4.6 & 6.1 & [+2.7, +6.5] & 32/9/1 & 0.000431 \\
\bottomrule
\end{tabular}
\caption{Aggregate gain of each critique method over each baseline, pooled across (model, dataset) points. Each point contributes its per-question mean accuracy difference in percentage points. The 95\% CI is $t$-based; the sign test reports the two-sided $p$-value for $H_0$ that positive and negative gains are equally likely (ties excluded).}
\label{tab:aggregate_gain}
\end{table}

To complement the per-dataset patterns above, \Cref{tab:aggregate_gain} reports aggregate gains pooled across all 42 (model, dataset) points. \EPCRIT{} and \EPSEMCRIT{} both significantly outperform \EPLABEL{}, with identical win rates of 32/42 and sign-test $p$-values of 0.000941 and 0.000431 respectively. Against the weaker \ZEROSHOT{} baseline, all three critique methods show significant gains, with \EPSEMCRIT{} positive on 38 of 42 points ($p < 10^{-7}$). These results demonstrate the overall utility of critique-based methods over both baselines.

Notably, \SEMCRIT{} alone does not reliably improve over \EPLABEL{} ($p=0.636$), confirming that semantic memory's value is primarily as a complement to episodic retrieval rather than as a standalone strategy. This is consistent with the roles we propose: episodic memory provides instance-specific reasoning, while semantic memory provides task-level guardrails that are most effective when combined with concrete examples.

\subsection{Critique-based Memory for Reasoning Models}
\label{sec:token_analysis}

\begin{table}[t]
\centering
\scriptsize
\setlength{\tabcolsep}{3pt}
\begin{tabular}{lccccccc}
\toprule
\makecell{Model and \\ Experiment} & \makecell{Multi-\\Cond.\\Ranking} & \makecell{NFCorpus} & \makecell{PubMed} & \makecell{Steam\\Pref} & \makecell{Book\\Pref} & \makecell{Anime\\Pref} & \makecell{Movie\\Pref} \\
\midrule
\multicolumn{8}{l}{\textbf{gpt-5-2025-08-07}} \\
\quad \EPLABEL & 3152 & 398 & \textbf{445} & 985 & 1127 & 1074 & 962 \\
\quad \EPCRIT{} & \underline{1683} & \underline{294} & \underline{502} & 784 & 632 & 792 & 758 \\
\quad \SEMCRIT{} & 1759 & 574 & 510 & \textbf{576} & \underline{579} & \textbf{638} & \textbf{581} \\
\quad \EPSEMCRIT{} & \textbf{1354} & \textbf{290} & 508 & \underline{694} & \textbf{568} & \underline{728} & \underline{734} \\
\hline
\multicolumn{8}{l}{\textbf{gpt-oss-20b}} \\
\quad \EPLABEL & 6357 & 504 & 260 & 1922 & 2328 & 2765 & 2718 \\
\quad \EPCRIT{} & 4790 & \underline{350} & \textbf{222} & \underline{836} & \underline{1079} & \underline{905} & \underline{781} \\
\quad \SEMCRIT{} & \textbf{3194} & 546 & 276 & \textbf{419} & \textbf{524} & \textbf{403} & \textbf{406} \\
\quad \EPSEMCRIT{} & \underline{4512} & \textbf{249} & \underline{232} & 864 & 1349 & 1147 & 862 \\
\hline
\multicolumn{8}{l}{\textbf{qwen3-vl-235b-a22b-thinking}} \\
\quad \EPLABEL & 6154 & 1345 & 868 & 1966 & 2099 & 2424 & 2466 \\
\quad \EPCRIT{} & \underline{4990} & \textbf{851} & \underline{813} & \underline{1691} & \underline{1574} & 1927 & \underline{1928} \\
\quad \SEMCRIT{} & 5151 & 1246 & \textbf{626} & \textbf{1170} & \textbf{1393} & \textbf{1331} & \textbf{1330} \\
\quad \EPSEMCRIT{} & \textbf{4741} & \underline{873} & 865 & 1814 & 1694 & \underline{1914} & 1984 \\
\bottomrule
\end{tabular}
\caption{Output token usage per question for thinking models across datasets and experiments. For each model and dataset, the lowest inference token count is \textbf{bolded} and the second-lowest is \underline{underlined}.}
\label{tab:token_usage_thinking}
\end{table}

Beyond accuracy improvements, we observe a surprising secondary benefit of critique-based memory for reasoning models: substantial reductions in inference-time computation. \Cref{tab:token_usage_thinking} reports the average number of output tokens (including thinking tokens) per question for our three reasoning models.

\paragraph{The presence of critiques reduces thinking tokens substantially compared to \EPLABEL{}}
Providing pre-computed critiques dramatically reduces the number of tokens reasoning models generate during inference. For example, compared to \EPLABEL{}, \EPCRIT{} reduces token usage by 67.3\% for GPT-oss-20B on Anime Pref. 
We hypothesize that these efficiency gains are caused by explicit critiques effectively substituting for reasoning the model would otherwise perform at inference time. The critiques provide conclusions the model can leverage immediately rather than derive independently. 

The critiques also appear to ground the thinking trajectories and help prevent them from getting derailed by spurious connections. For example, Qwen3-Thinking on preference datasets for \EPLABEL{} would frequently try to find connections between a user's numerical ID and the media the user preferred, despite there obviously being no connection. When augmented with critiques, we observed Qwen3-Thinking doing this much less frequently.

\paragraph{Critiques can prevent overthinking, improving performance.}
In some cases, this token reduction directly enables better performance in \Cref{tab:model_performance_reasoning}. On Multi-Condition Ranking, GPT-oss-20B achieves its best accuracy with \SEMCRIT{} - not because semantic memory provides superior signal, but because it prevents the model from exhausting its token budget through excessive deliberation. The percentage of answers that ran out of tokens dropped from 60\% with \EPLABEL{} to just 12\% with \SEMCRIT{}. This pattern repeats across other datasets and models: for example, on Movie Pref, GPT-oss-20B ran out of tokens 13.6\% of the time with \EPLABEL{} but only 0.4\% with \EPSEMCRIT{}. Likewise, for Qwen3-thinking on Multi-Condition Ranking, token-limited answers decreased from 40.5\% with \EPLABEL{} to 22.2\% with \EPSEMCRIT{}. 

GPT-5 almost never ran out of tokens, and as such did not experience this effect. It is possible GPT-5 would experience stronger gains with our CRIT methods compared to the \EPLABEL{} baseline if a stricter token budget was used, or if the reasoning effort parameter was set to high. See \Cref{tab:timeout_rate_thinking} for the full breakdown of token timeout rates across all models and experiments.

\paragraph{Semantic memory is particularly efficient.}
\SEMCRIT{} consistently produces the lowest token counts across most configurations. This aligns with its nature as a compressed, high-level summary rather than detailed per-example reasoning. While \SEMCRIT{} alone does not usually achieve the best accuracy, its efficiency makes it valuable in some circumstances.

\paragraph{Combination costs are not additive.}
The token cost of \EPSEMCRIT{} is substantially less than the sum of \EPCRIT{} and \SEMCRIT{} costs, and often only slightly higher than \EPCRIT{} alone. This indicates that the model does not naively process both memory types independently but rather integrates them efficiently. This means that there is little marginal cost to using the higher-performing \EPSEMCRIT{} strategy over \EPCRIT{}.

These efficiency gains have significant practical value. While generating memory incurs a one-time training cost (for example, averaging 5577 tokens per question for GPT-5 across datasets), this cost is amortized across all subsequent inference queries. In production scenarios where inference volume dominates total computation, pre-generating critiques not only improves accuracy but substantially reduces per-query costs and latency.  For a more detailed cost breakdown with break-even calculations, refer to \Cref{tab:usage}.

\subsection{Suggestibility}

\begin{table}[t]
\centering
\scriptsize
\setlength{\tabcolsep}{3pt}
\begin{tabular}{lccccccc}
\toprule
\makecell{Model and \\ Metric} & \makecell{Multi-\\Cond.\\Ranking} & \makecell{NFCorpus} & \makecell{PubMed} & \makecell{Steam\\Pref} & \makecell{Book\\Pref} & \makecell{Anime\\Pref} & \makecell{Movie\\Pref} \\
\midrule
\multicolumn{8}{l}{\textbf{gpt-4o-mini-2024-07-18}} \\
\quad Acc. (truth) & 94.8 & 93.6 & 100.0 & 100.0 & 100.0 & 100.0 & 100.0 \\
\quad Acc. (lying) & 9.2 & 74.4 & 1.6 & 0.0 & 0.0 & 0.0 & 0.0 \\
\quad Suggestibility & 85.6 & 19.2 & 98.4 & 100.0 & 100.0 & 100.0 & 100.0 \\
\hline
\multicolumn{8}{l}{\textbf{llama-4-scout}} \\
\quad Acc. (truth) & 64.0 & 90.0 & 95.2 & 92.4 & 82.0 & 81.6 & 87.2 \\
\quad Acc. (lying) & 36.0 & 71.6 & 27.2 & 11.6 & 16.4 & 20.8 & 15.2 \\
\quad Suggestibility & 28.0 & 18.4 & 68.0 & 80.8 & 65.6 & 60.8 & 72.0 \\
\hline
\multicolumn{8}{l}{\textbf{qwen3-235b-a22b-instruct-2507}} \\
\quad Acc. (truth) & 94.8 & 91.6 & 98.8 & 100.0 & 100.0 & 100.0 & 100.0 \\
\quad Acc. (lying) & 18.4 & 75.6 & 3.2 & 0.0 & 0.0 & 0.0 & 0.0 \\
\quad Suggestibility & 76.4 & 16.0 & 95.6 & 100.0 & 100.0 & 100.0 & 100.0 \\
\hline
\multicolumn{8}{l}{\textbf{gpt-5-2025-08-07}} \\
\quad Acc. (truth) & 100.0 & 98.4 & 100.0 & 99.2 & 100.0 & 100.0 & 100.0 \\
\quad Acc. (lying) & 26.0 & 61.6 & 0.4 & 1.2 & 0.0 & 0.0 & 0.0 \\
\quad Suggestibility & 74.0 & 36.8 & 99.6 & 98.0 & 100.0 & 100.0 & 100.0 \\
\hline
\multicolumn{8}{l}{\textbf{gpt-oss-20b}} \\
\quad Acc. (truth) & 90.4 & 95.6 & 98.8 & 99.6 & 100.0 & 99.6 & 100.0 \\
\quad Acc. (lying) & 49.2 & 73.2 & 2.0 & 0.0 & 0.4 & 0.4 & 0.0 \\
\quad Suggestibility & 41.2 & 22.4 & 96.8 & 99.6 & 99.6 & 99.2 & 100.0 \\
\hline
\multicolumn{8}{l}{\textbf{qwen3-vl-235b-a22b-thinking}} \\
\quad Acc. (truth) & 90.0 & 98.8 & 100.0 & 99.6 & 100.0 & 100.0 & 100.0 \\
\quad Acc. (lying) & 42.8 & 46.8 & 0.0 & 0.0 & 0.0 & 0.0 & 0.0 \\
\quad Suggestibility & 47.2 & 52.0 & 100.0 & 99.6 & 100.0 & 100.0 & 100.0 \\
\bottomrule
\end{tabular}
\caption{Suggestibility analysis across datasets. Acc. (truth) is accuracy when the agent provides truthful feedback, Acc. (lying) is accuracy when the agent provides deceptive feedback, and Suggestibility is the difference (truth - lying). Results on preference datasets are averaged across all users.}
\label{tab:suggestibility}
\end{table}

The variance in improvement across models raises a natural question: why do some models benefit more from critique-based memory augmentation than others? We hypothesize that a key factor is \textit{suggestibility}—the degree to which a model's predictions can be influenced by external reasoning provided in context. In memory-augmented agentic learning, it is crucial not only to generate high-quality critiques but also to ensure that the model is receptive to them. A model that rigidly adheres to its parametric knowledge, regardless of contextual information, will not benefit from memory augmentation.

To better quantify this phenomenon, we define a \textit{suggestibility} metric $S$, which captures the difference in an agent’s performance when given a best-effort critique versus when given an intentionally misleading one (generated by flipping the ground-truth label). Formally,
\begin{align*}
    S =\;& 
    \frac{1}{|D|} \sum_{x_i \in D} 
    \mathbb{1} \left[ \text{PA}(x_i \mid \text{CA}(x_i, y_i)) = y_i \right] - \\
    & 
    \frac{1}{|D|} \sum_{x_i \in D} 
    \mathbb{1} \left[ \text{PA}(x_i \mid \text{CA}(x_i, \neg y_i)) = y_i \right]
\end{align*}
where $\text{PA}$ denotes the performance agent, $\text{CA}$ refers to the critic agent, and $D$ is the evaluation dataset. Note that in real-world settings, the true label $y_i$ is not available to either $\text{PA}$ or $\text{CA}$; thus, this metric represents an idealized or “cheating” scenario, using artificially constructed best and adversarial insights for controlled experimentation.

\Cref{tab:suggestibility} reports suggestibility scores across all models and datasets.

\paragraph{Models are generally more suggestible on preference data than on fact-oriented data.} We hypothesize that this reflects a tension between knowledge encoded in a pretrained LLM’s parameters and the information provided by labeled examples. In fact-based domains, LLMs are likely more competent due to prior exposure to relevant facts during pretraining. A notable exception is the PubMed dataset, where the complexity of medical queries introduces enough ambiguity for critiques to meaningfully influence model outputs. In contrast, in preference-based domains, models cannot have learned individual user preferences—especially with anonymized users—so they lack parametric knowledge.

\paragraph{Suggestibility positively correlates with the benefit the model sees with memory augmentation.}
To assess whether suggestibility explains performance gains, we computed correlations between suggestibility and improvement across all 42 model-dataset combinations. When measuring improvement on \EPSEMCRIT{} relative to \ZEROSHOT{}, we observe a Spearman correlation coefficient of 0.402 with a p value of 0.008. This moderate positive correlation suggests that suggestibility partially explains why certain models benefit more from memory augmentation than others. For instance, Llama 4 Scout's low suggestibility scores (averaging 56.2\% across datasets) help account for its limited gains, while the high suggestibility of GPT-5 and Qwen3-Instruct aligns with their consistent high improvements over \ZEROSHOT{}.

\paragraph{However, suggestibility does not address gains of critique-based methods over few-shot learning.}
When measuring the improvement of \EPSEMCRIT{} relative to \EPLABEL{} - which already provides retrieved examples with correct labels - the correlation weakens substantially to 0.077 and is no longer statistically significant with a p value of 0.626. This suggests that while suggestibility partially explains a model's ability to benefit from external information generally, it does not account for the marginal benefit that structured critiques provide over raw input-output examples. The additional value of critique-based methods likely depends on factors beyond suggestibility such as whether the task contains generalizable patterns that critiques can articulate and whether the model can extract such patterns from examples in the critique-generation process.

\paragraph{Low suggestibility may impose a ceiling on possible gains.}
Llama 4 Scout exhibits the lowest average suggestibility across datasets (56.2\%) and shows the smallest improvements from critique-based methods, outperforming baselines on only 3 of 7 datasets. Even on preference data, where the model has no parametric knowledge of the user's preferences, the model frequently refused to return the correct answer. This may suggest that sufficient suggestibility may be a necessary but insufficient condition for benefiting from memory augmentation.

\paragraph{High suggestibility comes with safety concerns.}
While high suggestibility enables the benefits we observe in this work, it also raises concerns when models accept false information that contradicts their own knowledge. GPT-5's willingness to follow incorrect critiques on fact-based datasets (e.g., 74\% suggestibility on Multi-Condition Ranking, where it achieves 81.6\% zero-shot accuracy) suggests that even highly capable models may be overly deferential to contextual information. This reflects a tension in current alignment approaches: the same receptiveness that enables learning from feedback also creates vulnerabilities to adversarial or erroneous inputs. We leave deeper investigation of this tradeoff to future work.

%% file: tex/related.tex
\section{Related Work}
\label{sec:related}

\paragraph{Agentic Memory}
Much LLM-based research lately has been focused on agents, which use LLMs as the reasoning core of a larger compound system. While agents have shown significant improvements in performance across a variety of domains, a particularly difficult part of working with agents remains managing memory. Due to context length limitations and cost constraints, the agent's entire history cannot be included in every single message. Thus, memories must be encoded, processed, and retrieved only as necessary. Though there are many ways to do this, the most popular way is retrieval-based augmentation (RAG) which projects memories and user queries into a shared embedding space. Memories are then retrieved based on their similarity to the query in this space.

Memories can be as simple as direct copies of inputs and outputs, but frequently they are much more complex. Reflexion \citep{NEURIPS2023_1b44b878} regularly prompts the agent to self-reflect on its actions, and it is these reflections that are stored as memories. In Voyager \citep{Voyager}, memories are agent-created tools that may be re-used as needed. A-MEM \citep{xu2025amem} dynamically organizes memories into interconnected knowledge networks through adaptive tagging and linking, allowing the memory structure to evolve as new information is integrated.

A growing body of work draws on cognitive-architecture abstractions \citep{sumers2024cognitive,huet2025episodic} that separate \emph{episodic} traces of specific past experiences from \emph{semantic} knowledge generalized across them, and explores hierarchical organizations spanning these levels. G-Memory \citep{zhang2025gmemory} structures multi-agent interaction history into three tiers --- insight, query, and interaction graphs --- to support cross-trial knowledge transfer. H-MEM \citep{sun2025hmem} arranges memories across multiple semantic-abstraction layers with index-based routing for efficient retrieval in long-horizon agents, and MemInsight \citep{salama-etal-2025-meminsight} autonomously augments raw interactions with semantic attributes for more contextualized retrieval. MemoryBank \citep{Zhong_Guo_Gao_Ye_Wang_2024} similarly maintains long-term user memories with periodic summarization. Within this landscape, our framework adopts a deliberately simple two-tier structure --- instance-level episodic critiques and a single dataset-level semantic summary --- chosen for reproducibility and interpretability rather than for scaling to long-horizon multi-agent settings. This approach is appropriate for the relatively small datasets in this paper, but future work may benefit from exploring more complex hierarchical memory structures when adapting these techniques to larger datasets.

\paragraph{Fine-tuning}

Though general-purpose large language models can be very capable at certain tasks, they are limited by what is in their training data. To create useful models for niche use-cases, the most obvious approach is to fine-tune a foundation model on a new dataset. Using the principles of transfer learning, it is possible to greatly specialize models to specific tasks \cite{dodge2020finetuningpretrainedlanguagemodels}.

However, this is not without downsides. Though fine-tuning requires much less data than training foundation models from scratch, it still requires a very large amount of labeled data \cite{vieira2024datadatafinetuninglarge}, which may be difficult and expensive to obtain. Changing a model's weights often leads to "catastrophic forgetting," in which the fine-tuned model loses much of its general capabilities \cite{catastrophic_forgetting}. Fine-tuning is more computationally demanding than simple inference, and can be extremely expensive on larger models \cite{hu2022lora}. Lastly, fine-tuning requires access to the model's weights, meaning it cannot be done at all to closed-source models.

\paragraph{In-Context Learning}

Due to these issues with fine-tuning large language models, much research has been done investigating other learning mechanisms that instead treat the models as black boxes. These have come to be referred to generally as in-context learning strategies \citep{dong-etal-2024-survey}.

Largely, they all work on the same concept. LLMs are probabilistic models, and the trajectory of their output depends entirely on the input provided in their context window. By carefully adjusting what input are given to the model, it is possible to shift the output trajectory to something more favorable. This can take the form of presenting new information to the model, or simply changing how the question is asked.

In \citet{ZeroShot}, it was demonstrated that simply appending "Let's think step by step" to a question significantly improved performance across a variety of domains. Safety researchers have determined it is possible to counteract alignment fine-tuning efforts, or "jailbreak" models, through many different kinds of prompts, such as asking questions in code or presenting the question as a matter of life or death.

Beyond simply changing how questions are worded, it is also possible to improve performance by presenting new information. As demonstrated in \citet{FewShot}, presenting several examples of questions and the associated correct answers, or few-shot learning, is an extremely powerful technique to improve performance. 

Reflection-based in-context learning strategies are also becoming popular methods of specializing models to new datasets without relying on fine-tuning \citep{NEURIPS2023_1b44b878, yao2023react, NEURIPS2023_91edff07, yang2025thelight}. Instead of just presenting a few examples of correct question-answer pairs like in few-shot learning, the model makes decisions and then reasons over feedback it receives over the results of those decisions. By iteratively learning the effects of its actions, the model is able to learn how to make better decisions autonomously. However, this research has almost always been focused on giving the model feedback from a simulated environment \citep{Voyager, zhang2025spiralsymbolicllmplanning}, mimicking more traditional reinforcement-learning approaches. Relatively little research has been done on reflection/critique-based in-context learning using other kinds of feedback.

%% file: tex/conclusions.tex
\section{Discussion}

We observe consistent, significant gains  with self-critique strategies over baselines.
\Cref{tab:aggregate_gain} shows that \EPSEMCRIT{} improves over \EPLABEL{} on 32 of 42 (model, dataset) points. The remaining 10 non-improving cases are not randomly distributed; they cluster into three interpretable groups, each with an independent explanation.

\textbf{Llama 4 Scout} Four non-improving cases involve Llama 4 Scout (Multi-Condition Ranking, PubMed, Steam Pref, and Anime Pref). This is consistent with its low average suggestibility of 56.2\% (\Cref{tab:suggestibility}), which limits its ability to benefit from any in-context guidance, critiques included. Low suggestibility appears to impose a ceiling on gains from memory augmentation regardless of critique quality.

\textbf{Reasoning models on preference data} Four cases involve reasoning models on preference datasets: GPT-5 on Anime Pref and Movie Pref, and Qwen3-thinking on Anime Pref and Book Pref (the latter a tie). On these datasets \EPLABEL{} already yields substantial gains over \ZEROSHOT{} --- 8-20 absolute points across the affected model-dataset pairs --- and our token analysis in \Cref{sec:token_analysis} shows both models reduce thinking tokens substantially when critiques are provided. Taken together, this suggests these reasoning models perform critique-like reasoning internally over retrieved examples, making explicit critiques largely redundant for accuracy while still delivering the efficiency benefits documented in \Cref{sec:token_analysis}.

\textbf{PubMed} The remaining two cases are both on PubMed, for Qwen3-instruct and Qwen3-thinking, consistent with the pattern identified in \Cref{sec:main_results}: PubMed saw the least improvement across all models compared with both \ZEROSHOT{} and \EPLABEL{}. PubMed requires very specific factual knowledge that is rarely shared across questions, limiting how much critiques generated on training examples can transfer to test questions.

Each of the three groups points to a different boundary condition on the method: model suggestibility limitations, high token limits that permit internal critique-like reasoning, and domain structure that limits cross-instance transfer of critiques. None of the failures reflect random noise.

A second major pattern we observed was the consistent preference for episodic over semantic memory. This reflects a fundamental distinction between lazy and eager generalization strategies. Much like k-nearest neighbor methods, which often outperform regression-based approaches when training data is abundant, our lazy-learning episodic method has the advantage of not needing to commit to learning the full function over the entire domain. Instead, it learns only a local approximation to the function at the current query point, allowing for greater flexibility and specificity that may be lost during semantic abstraction.

Our analysis has centered on the accuracy of different agentic learning strategies, but design choices also impact computational cost and the ability to incorporate ongoing supervision. Semantic memory requires additional cost and complexity over episodic memory. At inference time, however, semantic memory offers more readily applicable knowledge, while episodic memory relies on retrieval quality. This trade-off suggests that the optimal strategy may depend on the size of the supervised dataset and the frequency of inference---semantic memory may be better suited for frequent inference under sparse supervision, while episodic memory may be preferable when supervision is abundant and retrieval is reliable.

An additional interesting direction for exploring model suggestibility is to disentangle how much a model's behavior changes due to genuinely incorporating supervised signals into its internal beliefs versus merely adapting its responses to please the user. In our empirical study, we observed that models exhibited higher suggestibility scores when critiques were attributed to the user, compared to when the same critiques were believed to originate from the model itself or another model. This suggests that the perceived source of feedback plays a significant role in how seriously the model treats the signal, opening up opportunities to better understand and guide belief formation in interactive learning systems.

\section{Conclusion}

In this work, we introduced a reflective framework for agent adaptation that leverages episodic and semantic memory to learn from labeled supervision without parameter updates. Our experiments demonstrate that by storing and retrieving label-driven critiques, LLMs can achieve significant performance gains. Our best-performing strategy, \EPSEMCRIT, yields an average improvement of 8.1 percentage points over zero-shot baselines and 4.6 points over RAG-style few-shot baselines.

Beyond accuracy, our analysis through the lens of suggestibility reveals that a model's capacity to utilize self-reflection is a function of its receptivity to external reasoning, rather than just its scale. Furthermore, we demonstrate a critical efficiency benefit: reasoning models can 'offload' internal computation to retrieved critiques, reducing thinking tokens by an average of 31.95\%.

This memory-driven approach offers a lightweight, interpretable, and computationally efficient alternative to fine-tuning. By treating model adaptation as a reflective process rather than a purely statistical one, we move closer to agents that can learn from their experiences in real-time, maintaining high performance across diverse and evolving tasks.

%% file: tex/appendix.tex
\appendix

\section{Ethics Statement}

This research on memory-augmented learning for large language model agents raises several important ethical considerations that we wish to acknowledge.

Though our suggestibility work was focused on how the model's instruction-following ability varied with dataset, this kind of approach could also be used to more efficiently jailbreak models to spread misinformation. Future work should be careful to avoid developing tools to improve the suggestibility of models to the point that they spread harmful misinformation.

We also recognize that improved adaptation capabilities may exacerbate existing biases in these agents. Because the insights are generated by the agent itself, even with feedback from the labeled data, it could cause the agent to reinforce its preconceptions about the world, which may perpetuate harmful stereotypes. Future work should explore safeguards to identify and mitigate such bias amplification.

\section{Implementation Details}
\subsection{RAG Implementation}
\label{app:rag_details}
We used blevlabs/stella\_en\_v5 as our encoder model and FAISS as our vector database. Similarity was based purely on the encodings of the questions in each dataset.

Though we did experiment with fine-tuning an encoder model to increase the separation of the classes in each dataset (for example, preferred vs not-preferred games for each Steam user) in embedding space, we did not see significant improvements in performance.

\subsection{Dataset: Additional Details}
\label{app:data}

\paragraph{Multi-Condition Ranking:} The original data source was presented in a free-form response format. However, for this task, we transformed it into a multiple-choice format for easier evaluation. The model now has to choose the correct answer from four unique orderings of the items.

\paragraph{NFCorpus:} The original NFCorpus data source associated each article with many papers with varying degrees of separation, which we transformed into this pairwise setup by choosing one paper at the closest and furthest level of separation possible for each paper. Sampled 500 shortest combinations of articles and papers to avoid context-length issues.

\section{Prompts}

\begin{promptbox}{Critique Generation} 
\small

\textbf{Question:} \{Question\}

\textbf{Initial Agent Response:} \{PA Initial Prediction\}

The correct answer is \{Ground Truth Answer\}. Explain why this is the correct answer, following the following JSON format\: 

\{

\quad correct\_answer: correct\_answer, 

\quad local\_reason: Specific reasons why this answer is correct in this particular case., 

\quad global\_reason: General reasons why this answer is correct that can be applied to other questions.

\}. 

Respond only with JSON.
\end{promptbox}

\begin{promptbox}{Semantic Memory Generation} 
\small
Your job is to summarize a set of self-critiques made by some agent as they perform different instances of their task.  For each instance you will be shown the output of the agent, followed by the critiques made by the agent after they were told the correct answer.  Distill those critiques into a helpful summary of advice to the agent, paying particular attention to instances where the agent outputs an incorrect answer. Produce your output in a form that can be used directly as instructions to the agent. You should summarize the key points in these critiques.  Be precise and concise.  Do not repeat yourself.

For example in train\_set:

\quad \{Question\} \{Answer\} \{Critique\} 

\end{promptbox}

\begin{promptbox}{Performance Agent with Semantic Memory} 
\small

\{Question\}

Here is some helpful advice that will help you make your decision: \{Summary\}
\end{promptbox}

\begin{promptbox}{Performance Agent with Episodic Memory} 
\small
For example in examples:

    \quad \textbf{Question:} \{Example Question\}
    
    \quad \textbf{Initial Agent Response:} \{PA Initial Prediction\}

    \quad (If \EPLABEL{}:) \textbf{Correct Answer:} \{correct\_answer\}

    \quad (If \EPCRIT{}:) \textbf{Reflection:} \{reflection\}

Here is your final question, make sure to learn from your past mistakes! \{Question\}
\end{promptbox}

\begin{promptbox}{Performance Agent with Episodic and Semantic Memory} 
\small
For example in examples:

    \quad \textbf{Question:} \{Example Question\}
    
    \quad \textbf{Initial Agent Response:} \{PA Initial Prediction\}

    \quad \textbf{Reflection:} \{reflection\}
    
Here is some additional advice to guide your response: \{Summary\} 

Here is your final question, make sure to learn from your past mistakes! \{Question\}

\end{promptbox}

\section{Examples}
\label{appendix_examples}

\begin{promptbox}{Example Semantic Memory

qwen3-vl-235b-a22b-thinking

Multi-Condition Ranking}
\begin{enumerate}[label=\textbf{\arabic*.}]

  \item \textbf{Interpret positional constraints correctly}
    \begin{itemize}
      \item ``Last from right'' means leftmost position in left-to-right list
      \item ``Last from left'' means rightmost position in left-to-right list
      \item These directional constraints override general priority ordering
    \end{itemize}

  \item \textbf{Categorize items by priority first}
    \begin{itemize}
      \item Identify high priority items (often with fixed position requirements)
      \item Identify medium priority items (typically based on specific attributes like ``related to X'' or ``born after Y'')
      \item Low priority items are all remaining items not classified as high/medium
    \end{itemize}

  \item \textbf{Apply sorting rules within priority groups}
    \begin{itemize}
      \item Low priority: sort by character count ascending (smallest to largest)
      \item Medium priority: sort by physical size (smallest to largest) or birthday (oldest to newest) as specified
      \item High priority: typically have fixed positions rather than being sorted among themselves
    \end{itemize}

  \item \textbf{Resolve ties systematically}
    \begin{itemize}
      \item When items have equal sorting values (e.g., same character count), use alphabetical order as tiebreaker
      \item Never assume equal values should maintain original order without explicit instruction
    \end{itemize}

  \item \textbf{Sequence priority groups correctly}
    \begin{itemize}
      \item The order of priority groups (high/medium/low) depends on specific problem constraints
      \item Verify which priority group should come first/last based on the conditions
      \item Fixed position constraints take precedence over general priority ordering
    \end{itemize}

  \item \textbf{Handle vacuous conditions properly}
    \begin{itemize}
      \item If no items meet a condition (e.g., no medium priority items), that condition doesn't affect the order
      \item Don't apply sorting rules to empty priority groups
    \end{itemize}

  \item \textbf{Verify all constraints sequentially}
    \begin{itemize}
      \item First satisfy fixed position requirements
      \item Then apply sorting rules to remaining items
      \item Always check that the final order satisfies all conditions simultaneously
    \end{itemize}

  \item \textbf{Prioritize explicit constraints over assumptions}
    \begin{itemize}
      \item When conditions conflict with your interpretation, re-examine the problem statement
      \item The correct answer must satisfy all explicit constraints, even if it contradicts your initial understanding of priority hierarchy
    \end{itemize}

\end{enumerate}
\end{promptbox}

\begin{promptbox}{Example Episodic Memory

qwen3-vl-235b-a22b-thinking

Multi-Condition Ranking}
\subsection*{Example 1:}
\textbf{Question:} Given the following conditions: ['Sort the items with a medium priority based on their height from the shortest to the tallest', 'Item ``bookshelf'' with a high priority should be the last from left', 'Sort the items with a low priority based on their character count from the smallest to largest'], sort the list of items ['bookshelf', 'water drop', 'chair', 'apartment', 'bottle'] from left to right. Choose the correct order from the provided options. Do not provide any explanation.\\
Option 1: ['bookshelf', 'apartment', 'water drop', 'bottle', 'chair']\\
Option 2: ['apartment', 'water drop', 'bottle', 'bookshelf', 'chair']\\
Option 3: ['water drop', 'bottle', 'chair', 'apartment', 'bookshelf']\\
Option 4: ['apartment', 'water drop', 'bottle', 'chair', 'bookshelf']\\
Respond only in JSON in the format \{``choice'': N\}, where N is 1 for Option 1, 2 for Option 2, and so on.

\textbf{Agent Response:} \{``choice'': 4\}

\textbf{Reflection:}
\begin{itemize}
  \item \textbf{Correct Answer:} 3
  \item \textbf{Local Reason:} The non-bookshelf items are sorted by height from shortest to tallest (water drop $<$ bottle $<$ chair $<$ apartment), satisfying the medium priority condition. `bookshelf' is correctly placed last as high priority. The low priority condition does not apply here as all non-high priority items are medium priority, making option 3 the only valid arrangement.
  \item \textbf{Global Reason:} When sorting with multiple priority levels, high priority items are fixed in their required position first. Remaining items are grouped by their priority level and sorted according to their respective criteria. The correct order combines these sorted groups while maintaining priority hierarchy, ensuring all specified conditions are met without contradiction.
\end{itemize}

\subsection*{Example 2:}
\textbf{Question:} Given the following conditions: ['Sort the items with a low priority based on their character count from the smallest to largest', 'Item ``bookshelf'' with a high priority should be the last from left', 'Sort the items with a medium priority based on their height from the shortest to the tallest'], sort the list of items ['ant', 'bookshelf', 'shoe', 'apartment', 'sofa'] from left to right. Choose the correct order from the provided options. Do not provide any explanation.\\
Option 1: ['bookshelf', 'shoe', 'sofa', 'apartment', 'ant']\\
Option 2: ['ant', 'shoe', 'sofa', 'apartment', 'bookshelf']\\
Option 3: ['sofa', 'ant', 'apartment', 'bookshelf', 'shoe']\\
Option 4: ['apartment', 'bookshelf', 'ant', 'sofa', 'shoe']\\
Respond only in JSON in the format \{``choice'': N\}, where N is 1 for Option 1, 2 for Option 2, and so on.

\textbf{Agent Response:} \{``choice'': 2\}

\textbf{Reflection:}
\begin{itemize}
  \item \textbf{Correct Answer:} 2
  \item \textbf{Local Reason:} The list must end with `bookshelf' (high priority). The remaining items are sorted such that low priority items (`ant', `shoe', `sofa') are ordered by character count (3, 4, 4), and medium priority (`apartment') follows. Option 2 correctly places `bookshelf' last and orders the others by character count for low priority items.
  \item \textbf{Global Reason:} When sorting with multiple priority levels, high priority items are fixed first (e.g., last position here), then lower priority groups are sorted by their specified criteria (e.g., character count or height). Correct answers must satisfy all constraints simultaneously, with higher priority rules taking precedence.
\end{itemize}

\subsection*{Example 3:}
...

\end{promptbox}

\section{Additional Results}

\subsection{Detailed Cost Analysis}

\begin{table}[t]
\centering
\scriptsize
\setlength{\tabcolsep}{3pt}
\begin{tabular}{lccccccc}
\toprule
\makecell{Model and \\ Metric} & \makecell{multi\\cond\\ranking} & \makecell{NFCorpus} & \makecell{PubMed} & \makecell{Steam\\Pref} & \makecell{Book\\Pref} & \makecell{Anime\\Pref} & \makecell{Movie\\Pref} \\
\midrule
\multicolumn{8}{l}{\textbf{gpt-5-2025-08-07}} \\
\quad Init Preds & 609.6k & 105.2k & 117.0k & 202.9k & 160.0k & 135.2k & 135.6k \\
\quad Critiques & 500.4k & 187.8k & 217.1k & 330.5k & 296.1k & 305.1k & 322.7k \\
\quad Summary & 3.3k & 3.7k & 5.4k & 29.0k & 27.3k & 31.7k & 33.6k \\
\quad BE: EP\_CRIT & 759 & 2.9k & never & 2.8k & 977 & 1.7k & 2.4k \\
\quad BE: SEM\_CRIT & 800 & never & never & 1.4k & 884 & 1.1k & 1.3k \\
\quad BE: EP+SEM\_CRIT & 620 & 2.8k & never & 1.9k & 866 & 1.4k & 2.2k \\
\hline
\multicolumn{8}{l}{\textbf{gpt-oss-20b}} \\
\quad Init Preds & 818.7k & 112.8k & 56.7k & 53.3k & 83.2k & 55.5k & 45.0k \\
\quad Critiques & 737.4k & 130.2k & 112.8k & 169.5k & 151.4k & 147.1k & 148.2k \\
\quad Summary & 1.5k & 893 & 2.1k & 14.3k & 15.5k & 14.3k & 13.7k \\
\quad BE: EP\_CRIT & 995 & 1.6k & 4.5k & 219 & 201 & 117 & 107 \\
\quad BE: SEM\_CRIT & 493 & never & never & 158 & 139 & 92 & 90 \\
\quad BE: EP+SEM\_CRIT & 845 & 956 & 6.2k & 225 & 256 & 135 & 112 \\
\hline
\multicolumn{8}{l}{\textbf{qwen3-vl-235b-a22b-thinking}} \\
\quad Init Preds & 1.30M & 195.8k & 121.8k & 211.3k & 278.2k & 230.1k & 238.9k \\
\quad Critiques & 1.08M & 219.1k & 147.1k & 179.5k & 187.0k & 199.1k & 141.5k \\
\quad Summary & 1.2k & 858 & 1.4k & 57.7k & 42.4k & 46.6k & 39.4k \\
\quad BE: EP\_CRIT & 2.0k & 842 & 5.0k & 1.6k & 967 & 958 & 781 \\
\quad BE: SEM\_CRIT & 2.4k & 4.2k & 1.1k & 564 & 719 & 436 & 370 \\
\quad BE: EP+SEM\_CRIT & 1.7k & 880 & 84.6k & 3.0k & 1.3k & 935 & 871 \\
\bottomrule
\end{tabular}
\caption{Token usage (completion tokens) per training stage and break-even test-question counts versus EP\_LABEL. Init Preds / Critiques / Summary are total train-set costs. All preference data numbers are summed across users, so all columns represent the cost across a training size of 250 questions. Semantic memory costs will be proportionally higher for preference data because it has to be generated separately for all ten users.}
\label{tab:usage}
\end{table}

\Cref{tab:usage} reports the total completion tokens used during offline memory generation (initial predictions, critiques, and semantic summary), along with the number of test queries at which each critique strategy amortizes its offline cost relative to \EPLABEL{}. Semantic summary generation is extremely cheap in all cases because it requires only a single LLM call (per-user) over the already-generated critiques; the dominant upfront costs are initial predictions and critique generation. This means that the accuracy gains we demonstrate for \EPSEMCRIT{} over \EPCRIT{} in the main paper come at essentially no additional memory-generation cost.

Break-even speed varies substantially by dataset type. Preference datasets break even quickly, sometimes within double-digit numbers of queries, while knowledge-intensive datasets require more queries to amortize, with PubMed being the highest by a significant margin. Notably, despite Multi-Condition Ranking being by far the most expensive to generate memories for, it demonstrates such significant efficiency gains at inference time that it breaks even within a few hundred queries. The ``never'' entries --- where a strategy does not break even --- occur in a small number of model-dataset-strategy combinations, primarily \SEMCRIT{} on NFCorpus and PubMed. In these cases the critique method uses more inference tokens than \EPLABEL{}, so the upfront generation cost is never amortized through token savings. However, we still observe significant accuracy gains over \EPLABEL{} on these datasets (for example, on NFCorpus \EPSEMCRIT{} improves over \EPLABEL{} for all six models), indicating that the accuracy and efficiency benefits of critique memory are complementary but do not always co-occur. Overall, for the majority of model-dataset combinations, the offline generation cost is amortized within a modest number of queries, making the approach practical for most production deployments.

\subsection{Training Data Size Ablation}

\begin{table}[t]
\centering
\scriptsize
\setlength{\tabcolsep}{3pt}
\begin{tabular}{lcccc | cccc}
\toprule
\makecell{Model and \\ Experiment} & \multicolumn{4}{c|}{\makecell{Multi-\\Cond.\\Ranking}} & \multicolumn{4}{c}{\makecell{Steam\\Pref}} \\
 & 25\% & 50\% & 75\% & 100\% & 25\% & 50\% & 75\% & 100\% \\
\midrule
\multicolumn{9}{l}{\textbf{gpt-4o-mini-2024-07-18}} \\
\quad \EPCRIT{} & 34.8 & \textbf{35.6} & 35.6 & 34.8 & 67.2 & 65.6 & 69.2 & \textbf{71.2} \\
\quad \EPSEMCRIT{} & 35.2 & \textbf{36.8} & 34.4 & 36.8 & 69.2 & 68.4 & 71.2 & \textbf{73.6} \\
\hline
\multicolumn{9}{l}{\textbf{gpt-5-2025-08-07}} \\
\quad \EPCRIT{} & \textbf{99.2} & 97.2 & 98.0 & 96.8 & 66.0 & 62.8 & 66.4 & \textbf{69.2} \\
\quad \EPSEMCRIT{} & 98.8 & 98.4 & 98.8 & \textbf{99.2} & 66.4 & 64.4 & 66.4 & \textbf{68.8} \\
\hline
\multicolumn{9}{l}{\textbf{gpt-oss-20b}} \\
\quad \EPCRIT{} & 52.8 & 52.4 & 54.0 & \textbf{57.6} & 59.2 & 57.6 & 60.4 & \textbf{63.2} \\
\quad \EPSEMCRIT{} & 57.6 & \textbf{58.8} & 55.2 & 58.8 & 59.6 & 58.0 & 60.8 & \textbf{62.8} \\
\bottomrule
\end{tabular}
\caption{Train-size ablation showing accuracy  of \EPCRIT{} and \EPSEMCRIT{} at different training-set fractions. \textbf{Bold} marks the best fraction per (model, dataset, experiment).}
\label{tab:train_size_ablation}
\end{table}

Our approach is naturally suited to online deployment. Critique generation for each training example is fully independent, so new episodic memories can be generated and added to the vector database in parallel as new labeled data arrives without regenerating existing memories, and semantic memory needs only periodic regeneration as the critique pool grows. To characterize how performance scales as labeled data accumulates, we evaluated \EPCRIT{} and \EPSEMCRIT{} at 25\%, 50\%, 75\%, and 100\% of the training data on a subset of models and datasets, where each larger split is a strict superset of the smaller ones. Because episodic memory is stateless and semantic memory is regenerated from the full critique pool at each stage, this is equivalent to an online agent that accumulates labeled examples over time and periodically refreshes its memory in batches.

As shown in \Cref{tab:train_size_ablation}, both \EPCRIT{} and \EPSEMCRIT{} show clear, largely monotonic improvement on Steam Preference as data accumulates across all three models, consistent with what we would expect from an online system that improves with experience. Multi-Condition Ranking shows a general upward trend but with more variance across splits. We hypothesize that embedding-based retrieval is less reliable on this dataset: Multi-Condition Ranking requires reasoning about relational properties between items rather than intrinsic item attributes, which may mean that the most useful critiques are not the most embedding-similar. As the pool grows, more superficially similar but less informative examples could crowd out the most useful ones. While the embedding-retrieved episodic memories remain clearly helpful over zero-shot, this suggests that alternative retrieval strategies such as diversity-aware or task-adapted retrieval are a promising direction for future work on tasks with this property. A second notable pattern is that on Multi-Condition Ranking, \EPSEMCRIT{} at 25\% of the training data always matches or exceeds full-data \EPCRIT{}, suggesting that semantic memory provides useful task-level signals even from small critique pools --- a property that would be especially valuable in early-stage online deployment when few labeled examples are available.

\subsection{Critique Component Ablation}

\begin{table}[t]
\centering
\scriptsize
\setlength{\tabcolsep}{3pt}
\begin{tabular}{lccccccc}
\toprule
\makecell{Model and \\ Variant} & \makecell{Multi-\\Cond.\\Ranking} & \makecell{NFCorpus} & \makecell{PubMed} & \makecell{Steam\\Pref} & \makecell{Book\\Pref} & \makecell{Anime\\Pref} & \makecell{Movie\\Pref} \\
\midrule
\multicolumn{8}{l}{\textbf{gpt-4o-mini-2024-07-18}} \\
\quad Local & \textbf{35.6} & \textbf{89.6} & 64.8 & \textbf{71.6} & 64.8 & 60.4 & \textbf{59.6} \\
\quad Global & 34.4 & 86.0 & \textbf{65.2} & 64.4 & 60.4 & 58.4 & 50.4 \\
\quad Full & 33.6 & 88.8 & 63.2 & \textbf{71.6} & \textbf{66.4} & \textbf{62.8} & 58.8 \\
\hline
\multicolumn{8}{l}{\textbf{gpt-5-2025-08-07}} \\
\quad Local & 97.2 & 97.6 & \textbf{72.8} & 68.8 & \textbf{70.0} & \textbf{75.6} & 64.4 \\
\quad Global & 97.6 & 97.6 & 68.8 & 66.0 & 67.6 & 67.6 & 63.2 \\
\quad Full & \textbf{98.0} & \textbf{98.4} & 70.8 & \textbf{69.2} & 69.2 & 71.6 & \textbf{67.2} \\
\hline
\multicolumn{8}{l}{\textbf{gpt-oss-20b}} \\
\quad Local & \textbf{59.2} & 94.8 & 66.4 & 61.6 & \textbf{65.2} & \textbf{64.0} & \textbf{66.4} \\
\quad Global & 44.4 & 94.8 & 66.0 & 63.6 & 60.0 & 60.0 & 58.4 \\
\quad Full & 56.4 & \textbf{95.2} & \textbf{67.2} & \textbf{64.8} & 60.4 & 60.8 & 63.6 \\
\bottomrule
\end{tabular}
\caption{Critique-parts ablation showing \EPCRIT{} accuracy when critiques contain only the local reason, only the global reason, or both (full). \textbf{Bold} marks the best variant per (model, dataset).}
\label{tab:critique_parts_ablation}
\end{table}

\Cref{tab:critique_parts_ablation} reports the accuracy of \EPCRIT{} when the critique attached to each retrieved example contains only the local reason, only the global reason, or the full critique (note, full is an independent trial from the accuracies reported in \Cref{tab:model_performance} and \Cref{tab:model_performance_reasoning}). A clear, model-dependent pattern emerges in how the two components are utilized. GPT-4o-mini and GPT-oss tend to benefit most from local rationales: the local-only variant is the best or tied-for-best on four of seven datasets for each, and the gap to global-only can be large --- nearly 15 points on Multi-Condition Ranking for GPT-oss (59.2 vs.\ 44.4) and roughly 9 points on Movie Preference for GPT-4o-mini (59.6 vs.\ 50.4). The stronger GPT-5 shows the opposite tilt: the full critique is best on four of seven datasets, and the gap between local-only and global-only is much smaller. This is consistent with a more capable model being better positioned to extract value from task-level patterns and to integrate them with per-instance rationales rather than relying on either signal alone.

Two broader observations follow. First, the global-only variant wins outright in only one of the 21 (model, dataset) cells (GPT-4o-mini on PubMed), indicating that instance-specific reasoning is the most consistently load-bearing component across the models and tasks we evaluate. Second, despite this, global reflections contribute meaningful complementary information once paired with local context --- most strikingly for GPT-5, but also for GPT-4o-mini on Book and Anime Preference and for GPT-oss on NFCorpus, PubMed, and Steam Preference, where the full variant matches or surpasses the local-only variant. Combining both components therefore provides broader coverage across the settings we evaluate than committing to a single signal type.

\subsection{Varying K}
\label{varying_k}

\begin{table}[htbp]
\scriptsize
\setlength{\tabcolsep}{3pt}
\centering
\begin{tabular}{lccccccc}
\toprule
\makecell{Model and \\ Experiment} & \makecell{Multi-\\Cond.\\Ranking} & \makecell{NFCorpus} & \makecell{PubMed} & \makecell{Steam\\Pref} & \makecell{Book\\Pref} & \makecell{Anime\\Pref} & \makecell{Movie\\Pref} \\
\midrule
\multicolumn{8}{l}{\EPCRIT{}} \\
\quad $K=1$ & 35.2 & 82.4 & 63.2 & 65.2 & 62.0 & 59.2 & 57.2 \\
\quad $K=3$ & 35.2 & 86.0 & 62.8 & 69.6 & 63.2 & 61.2 & 56.8 \\
\quad $K=5$ & 34.8 & 89.6 & 63.2 & {71.2} & 64.8 & 62.0 & {58.4} \\
\quad $K=10$ & {37.6} & {90.4} & {66.4} & 70.4 & {66.8} & {64.0} & 57.6 \\
\bottomrule
\end{tabular}
\caption{Effect on accuracy of varying $K$, the number of examples used in episodic memory, with gpt-4o-mini as the base model.}
\label{tab:varying_k}
\end{table}

In \Cref{tab:varying_k}, we observe that varying $K$, the number of examples included from the episodic memory module, can have significant effects on the accuracy of the overall performance agent, with an up to 8\% absolute difference in accuracy between different K values for the same setup. $K=5$ and $K=10$ both have similar results, and are on average better than $K=1$ or $K=3$. Because the results for $K=10$ were not significantly higher than $K=5$, we elected to use the simpler method and use $K=5$ for all episodic and episodic+semantic memory experiments in this paper.

\subsection{Model Token Timeout Rates}

\begin{table}[t]
\centering
\scriptsize
\setlength{\tabcolsep}{3pt}
\begin{tabular}{lccccccc}
\toprule
\makecell{Model and \\ Experiment} & \makecell{Multi-\\Cond.\\Ranking} & \makecell{NFCorpus} & \makecell{PubMed} & \makecell{Steam\\Pref} & \makecell{Book\\Pref} & \makecell{Anime\\Pref} & \makecell{Movie\\Pref} \\
\midrule
\multicolumn{8}{l}{\textbf{gpt-5-2025-08-07}} \\
\quad \textit{Train} & 1.2\ & 0.0\ & 0.0\ & 0.0\ & 0.0\ & 0.0\ & 0.0\ \\
\quad \ZEROSHOT{} & 2.0 & 0.0 & 0.0 & 0.0 & 0.0 & 0.0 & 0.0 \\
\quad \EPLABEL & 4.0 & 0.0 & 0.0 & 0.0 & 0.0 & 0.0 & 0.0 \\
\quad \EPCRIT{} & 1.2 & 0.0 & 0.0 & 0.0 & 0.0 & 0.0 & 0.0 \\
\quad \SEMCRIT{} & 0.4 & 0.0 & 0.0 & 0.0 & 0.0 & 0.0 & 0.0 \\
\quad \EPSEMCRIT{} & 0.4 & 0.0 & 0.0 & 0.0 & 0.0 & 0.0 & 0.0 \\
\hline
\multicolumn{8}{l}{\textbf{gpt-oss-20b}} \\
\quad \textit{Train} & 9.5\ & 0.5\ & 0.0\ & 0.1\ & 0.1\ & 0.0\ & 0.0\ \\
\quad \ZEROSHOT{} & 14.4 & 0.8 & 0.0 & 0.4 & 0.0 & 0.0 & 0.0 \\
\quad \EPLABEL & 60.0 & 1.2 & 0.0 & 6.8 & 8.4 & 12.0 & 13.6 \\
\quad \EPCRIT{} & 36.8 & 0.4 & 0.4 & 3.2 & 4.4 & 2.0 & 0.4 \\
\quad \SEMCRIT{} & 12.0 & 1.2 & 0.0 & 0.0 & 0.8 & 0.0 & 0.0 \\
\quad \EPSEMCRIT{} & 35.2 & 0.0 & 0.0 & 1.2 & 3.6 & 3.2 & 0.4 \\
\hline
\multicolumn{8}{l}{\textbf{qwen3-vl-235b-a22b-thinking}} \\
\quad \textit{Train} & 8.7\ & 0.1\ & 0.0\ & 0.0\ & 0.0\ & 0.0\ & 0.0\ \\
\quad \ZEROSHOT{} & 25.1 & 0.0 & 0.0 & 0.0 & 0.0 & 0.0 & 0.0 \\
\quad \EPLABEL & 40.5 & 1.2 & 0.0 & 0.0 & 0.0 & 0.4 & 0.0 \\
\quad \EPCRIT{} & 26.2 & 0.0 & 0.0 & 0.0 & 0.0 & 0.0 & 0.0 \\
\quad \SEMCRIT{} & 29.6 & 1.2 & 0.0 & 0.0 & 0.0 & 0.0 & 0.0 \\
\quad \EPSEMCRIT{} & 22.2 & 0.0 & 0.0 & 0.0 & 0.0 & 0.0 & 0.0 \\
\bottomrule
\end{tabular}
\caption{Timeout rate -  the percentage of responses where output tokens reached the token limit - for thinking models across datasets and experiments.}
\label{tab:timeout_rate_thinking}
\end{table}

In \Cref{tab:timeout_rate_thinking} we report token timeout rates by model and dataset. We include only reasoning models, as none of the non-reasoning models timed out on any question. The token limit was set to 8,192 tokens, counting both the internal reasoning trace and the final output.

Timeout rates vary substantially across both models and datasets. All three models timed out most frequently on Multi-Condition Ranking, by a wide margin. This is likely because these questions demand the greatest amount of explicit reasoning: rather than relying on external knowledge, they require the model to execute a sequence of logical steps to sort items according to multiple criteria. Given this structure, it is unsurprising that this dataset exhibits the highest timeout rates.

Even on the multi-condition ranking task, GPT-5 almost never timed out, suggesting substantially more efficient use of reasoning tokens. In contrast, GPT-oss experienced more timeouts than Qwen3-Thinking on the preference datasets, despite using far fewer tokens per question on average (see \Cref{tab:token_usage_thinking}). This apparent contradiction is explained by the much higher variance in GPT-oss’s response lengths. While GPT-oss often produces relatively concise outputs, it is also more prone to generating unusually long reasoning traces that exceed the token limit. On manual inspection, GPT-oss appeared to be particularly prone to getting stuck in loops during its reasoning trace.

\subsection{Per-User Preference Data Results}

When presenting the results for the preference datasets elsewhere in the paper, the results are averaged across 10 users per dataset. This is done because different users have very different preference patterns, and some users are more consistent than others. Sampling across many users gives us a better look at the models true performance at this task for the general population.
In \Cref{tab:per_user_steam}, \Cref{tab:per_user_book}, \Cref{tab:per_user_anime}, and \Cref{tab:per_user_movie} we present the results separated out by user, along with the average and standard deviation across all ten users.

\begin{table*}[ht]
\centering
\begin{subtable}[t]{0.49\textwidth}
\centering
\scriptsize
\setlength{\tabcolsep}{2.5pt}
\begin{tabular}{lcccccccccccc}
\toprule
\makecell{Model and \\ Experiment} & U.1 & U.2 & U.3 & U.4 & U.5 & U.6 & U.7 & U.8 & U.9 & U.10 & Avg & Std \\
\midrule
\multicolumn{13}{l}{\textbf{gpt-4o-mini-2024-07-18}} \\
\quad \ZEROSHOT{} & 52 & 68 & \textbf{80} & 40 & \underline{64} & 72 & 24 & 48 & 64 & 72 & 58 & 16 \\
\quad \EPLABEL & 56 & 68 & 60 & 48 & 52 & 64 & 36 & \textbf{80} & 68 & \underline{80} & 61 & 13 \\
\quad \EPCRIT{} & \underline{76} & \underline{80} & \underline{80} & \textbf{60} & \textbf{68} & 76 & \textbf{64} & 60 & \textbf{76} & 72 & 71 & 7 \\
\quad \SEMCRIT{} & 72 & 68 & 68 & 48 & 56 & \textbf{80} & 36 & 48 & 52 & 76 & 60 & 14 \\
\quad \EPSEMCRIT{} & \textbf{80} & \textbf{84} & 76 & \underline{60} & 64 & \underline{80} & \underline{60} & \underline{72} & \underline{72} & \textbf{88} & 74 & 9 \\
\hline
\multicolumn{13}{l}{\textbf{gpt-5-2025-08-07}} \\
\quad \ZEROSHOT{} & 56 & \underline{60} & 60 & 56 & 44 & 60 & 44 & 56 & 52 & 48 & 54 & 6 \\
\quad \EPLABEL & 60 & 44 & \underline{80} & 64 & 56 & \textbf{68} & \textbf{80} & 60 & \underline{64} & 68 & 64 & 10 \\
\quad \EPCRIT{} & \textbf{80} & \textbf{72} & 76 & \textbf{72} & 60 & \underline{64} & \underline{76} & 64 & 56 & \textbf{72} & 69 & 7 \\
\quad \SEMCRIT{} & 64 & 60 & 68 & 60 & \underline{64} & 64 & 48 & \textbf{76} & \textbf{72} & 68 & 64 & 7 \\
\quad \EPSEMCRIT{} & \underline{72} & 60 & \textbf{84} & \underline{68} & \textbf{72} & 56 & 76 & \underline{68} & 60 & \underline{72} & 69 & 8 \\
\hline
\multicolumn{13}{l}{\textbf{gpt-oss-20b}} \\
\quad \ZEROSHOT{} & 52 & 52 & 44 & \underline{64} & 40 & 64 & 36 & \underline{52} & \textbf{72} & 48 & 52 & 11 \\
\quad \EPLABEL & 56 & 52 & 60 & 60 & 44 & 56 & \textbf{64} & \textbf{60} & 44 & \textbf{76} & 57 & 9 \\
\quad \EPCRIT{} & \textbf{76} & \textbf{72} & \textbf{68} & 56 & \underline{60} & \underline{76} & 56 & 44 & 56 & \underline{68} & 63 & 10 \\
\quad \SEMCRIT{} & 56 & 56 & 48 & \textbf{76} & 36 & \textbf{80} & 52 & 44 & \underline{68} & 56 & 57 & 13 \\
\quad \EPSEMCRIT{} & \underline{68} & \underline{68} & \underline{64} & 56 & \textbf{64} & 76 & \underline{60} & 52 & 52 & 68 & 63 & 7 \\
\hline
\multicolumn{13}{l}{\textbf{llama-4-scout}} \\
\quad \ZEROSHOT{} & 60 & 72 & \textbf{72} & 48 & \underline{48} & 56 & 32 & 44 & 56 & \underline{76} & 56 & 13 \\
\quad \EPLABEL & \textbf{72} & \textbf{76} & 60 & \textbf{64} & \textbf{52} & \textbf{80} & \textbf{44} & \textbf{68} & \textbf{72} & \textbf{80} & 67 & 11 \\
\quad \EPCRIT{} & 68 & 72 & 60 & \underline{64} & 48 & 76 & 36 & 56 & 68 & 76 & 62 & 12 \\
\quad \SEMCRIT{} & \underline{72} & 72 & 60 & 52 & 48 & \underline{80} & \underline{44} & \underline{64} & 68 & 72 & 63 & 11 \\
\quad \EPSEMCRIT{} & 72 & \underline{76} & \underline{64} & 60 & 40 & 80 & 40 & 60 & \underline{72} & 72 & 64 & 13 \\
\hline
\multicolumn{13}{l}{\textbf{qwen3-235b-a22b-instruct-2507}} \\
\quad \ZEROSHOT{} & 52 & 48 & 68 & 44 & 56 & \underline{72} & 32 & \textbf{64} & \underline{60} & 52 & 55 & 11 \\
\quad \EPLABEL & 60 & 44 & 64 & 44 & 40 & 52 & 52 & \underline{64} & \textbf{72} & 64 & 56 & 10 \\
\quad \EPCRIT{} & \textbf{80} & 60 & \textbf{88} & \textbf{68} & \textbf{60} & 64 & \underline{72} & 60 & 60 & 72 & 68 & 9 \\
\quad \SEMCRIT{} & 52 & \textbf{72} & \underline{84} & 52 & \underline{60} & \textbf{80} & 56 & 52 & 56 & \textbf{76} & 64 & 12 \\
\quad \EPSEMCRIT{} & \underline{80} & \underline{64} & 76 & \underline{68} & 60 & 64 & \textbf{80} & 56 & 60 & \underline{76} & 68 & 8 \\
\hline
\multicolumn{13}{l}{\textbf{qwen3-vl-235b-a22b-thinking}} \\
\quad \ZEROSHOT{} & \textbf{68} & \textbf{80} & 60 & \underline{56} & 56 & \underline{60} & 32 & 48 & \textbf{68} & 60 & 59 & 12 \\
\quad \EPLABEL & 48 & 46 & \textbf{68} & 56 & 56 & 52 & 64 & 48 & \underline{60} & 68 & 57 & 8 \\
\quad \EPCRIT{} & 60 & 72 & 64 & \textbf{60} & \textbf{72} & 52 & \textbf{76} & \textbf{56} & 60 & \underline{72} & 64 & 8 \\
\quad \SEMCRIT{} & 40 & 52 & \underline{68} & 44 & \underline{64} & \textbf{64} & 44 & \underline{56} & 60 & \textbf{80} & 57 & 12 \\
\quad \EPSEMCRIT{} & \underline{64} & \underline{76} & 68 & 56 & 60 & 44 & \underline{76} & 52 & 56 & 68 & 62 & 10 \\
\bottomrule
\end{tabular}
\caption{Per-user accuracy for Steam Preference.}
\label{tab:per_user_steam}
\end{subtable}
\hfill
\begin{subtable}[t]{0.49\textwidth}
\centering
\scriptsize
\setlength{\tabcolsep}{2.5pt}
\begin{tabular}{lcccccccccccc}
\toprule
\makecell{Model and \\ Experiment} & U.1 & U.2 & U.3 & U.4 & U.5 & U.6 & U.7 & U.8 & U.9 & U.10 & Avg & Std \\
\midrule
\multicolumn{13}{l}{\textbf{gpt-4o-mini-2024-07-18}} \\
\quad \ZEROSHOT{} & \underline{56} & \textbf{80} & 64 & 40 & 60 & 40 & 60 & 48 & 56 & 52 & 56 & 11 \\
\quad \EPLABEL & \textbf{60} & 68 & 60 & \underline{52} & \textbf{64} & 60 & 56 & 52 & 52 & 56 & 58 & 5 \\
\quad \EPCRIT{} & 56 & 72 & \underline{72} & \textbf{60} & 56 & \textbf{64} & \underline{68} & \underline{56} & \textbf{72} & \underline{72} & 65 & 7 \\
\quad \SEMCRIT{} & 52 & \underline{76} & 68 & 36 & \underline{64} & 56 & 68 & 48 & 56 & 64 & 59 & 11 \\
\quad \EPSEMCRIT{} & 56 & 72 & \textbf{80} & 44 & 60 & \underline{64} & \textbf{76} & \textbf{64} & \underline{72} & \textbf{80} & 67 & 11 \\
\hline
\multicolumn{13}{l}{\textbf{gpt-5-2025-08-07}} \\
\quad \ZEROSHOT{} & 48 & 64 & 56 & 56 & 52 & 32 & 56 & 52 & 52 & 52 & 52 & 8 \\
\quad \EPLABEL & \textbf{76} & \textbf{68} & \underline{80} & 68 & \underline{64} & 64 & \textbf{72} & \textbf{72} & \underline{72} & 76 & 71 & 5 \\
\quad \EPCRIT{} & 64 & \underline{68} & 64 & \textbf{76} & \textbf{68} & 64 & \underline{72} & 64 & \textbf{76} & \textbf{84} & 70 & 7 \\
\quad \SEMCRIT{} & 68 & 68 & 80 & 64 & 64 & \underline{68} & 60 & \underline{72} & 64 & 48 & 66 & 8 \\
\quad \EPSEMCRIT{} & \underline{76} & 68 & \textbf{88} & \underline{76} & 64 & \textbf{76} & 72 & 64 & 68 & \underline{80} & 73 & 7 \\
\hline
\multicolumn{13}{l}{\textbf{gpt-oss-20b}} \\
\quad \ZEROSHOT{} & \textbf{60} & 64 & 56 & \underline{36} & 52 & \underline{64} & \underline{64} & 52 & 52 & \textbf{64} & 56 & 8 \\
\quad \EPLABEL & \underline{56} & 68 & \textbf{80} & 28 & 36 & \textbf{72} & 52 & 52 & 56 & 56 & 56 & 15 \\
\quad \EPCRIT{} & 56 & \underline{72} & \underline{72} & 28 & 48 & 60 & 56 & \textbf{60} & \textbf{76} & 56 & 58 & 13 \\
\quad \SEMCRIT{} & 56 & \textbf{80} & 64 & \textbf{44} & \textbf{60} & 60 & 56 & \underline{56} & \underline{72} & 52 & 60 & 10 \\
\quad \EPSEMCRIT{} & 52 & 72 & 72 & 36 & \underline{56} & 56 & \textbf{68} & 52 & 72 & \underline{60} & 60 & 11 \\
\hline
\multicolumn{13}{l}{\textbf{llama-4-scout}} \\
\quad \ZEROSHOT{} & \textbf{56} & 72 & \textbf{68} & \textbf{44} & 52 & 52 & \underline{60} & 52 & 48 & \textbf{64} & 57 & 9 \\
\quad \EPLABEL & 48 & \underline{84} & 48 & 28 & 52 & 52 & 52 & \textbf{64} & 52 & \underline{64} & 54 & 14 \\
\quad \EPCRIT{} & 52 & 76 & \underline{68} & \underline{44} & \textbf{64} & \underline{56} & 60 & \underline{60} & \underline{68} & 60 & 61 & 9 \\
\quad \SEMCRIT{} & \underline{56} & 84 & 68 & 44 & \underline{60} & 44 & \textbf{68} & 44 & 68 & 56 & 59 & 12 \\
\quad \EPSEMCRIT{} & 56 & \textbf{96} & 56 & 44 & 60 & \textbf{60} & 60 & 60 & \textbf{76} & 48 & 62 & 14 \\
\hline
\multicolumn{13}{l}{\textbf{qwen3-235b-a22b-instruct-2507}} \\
\quad \ZEROSHOT{} & 52 & 68 & 56 & 44 & 60 & 40 & \textbf{76} & 52 & 52 & 48 & 55 & 10 \\
\quad \EPLABEL & 60 & 64 & 56 & 52 & 56 & \underline{60} & \underline{68} & 40 & 64 & \textbf{68} & 59 & 8 \\
\quad \EPCRIT{} & 60 & 64 & 68 & \underline{60} & 60 & \textbf{68} & 68 & \textbf{60} & \textbf{72} & 56 & 64 & 5 \\
\quad \SEMCRIT{} & \textbf{68} & \underline{72} & \textbf{80} & 52 & \underline{68} & 36 & 60 & \underline{60} & 48 & 60 & 60 & 12 \\
\quad \EPSEMCRIT{} & \underline{64} & \textbf{76} & \underline{76} & \textbf{68} & \textbf{76} & 56 & 64 & 52 & \underline{68} & \underline{64} & 66 & 8 \\
\hline
\multicolumn{13}{l}{\textbf{qwen3-vl-235b-a22b-thinking}} \\
\quad \ZEROSHOT{} & 44 & 72 & 48 & 32 & \underline{64} & 56 & 52 & \textbf{64} & 56 & 48 & 54 & 11 \\
\quad \EPLABEL & \textbf{64} & 72 & \underline{80} & 48 & 52 & \textbf{72} & 52 & 36 & \textbf{72} & \textbf{72} & 62 & 13 \\
\quad \EPCRIT{} & 52 & \textbf{76} & 64 & 44 & 56 & \underline{60} & \textbf{68} & 44 & 68 & 44 & 58 & 11 \\
\quad \SEMCRIT{} & 40 & 72 & \textbf{84} & \textbf{60} & \textbf{76} & 48 & \underline{64} & 36 & 60 & 48 & 59 & 15 \\
\quad \EPSEMCRIT{} & \underline{60} & \underline{76} & 76 & \underline{56} & 56 & 60 & 60 & \underline{48} & \underline{72} & \underline{56} & 62 & 9 \\
\bottomrule
\end{tabular}
\caption{Per-user accuracy for Book Preference.}
\label{tab:per_user_book}
\end{subtable}
\caption{For each model and user, the highest score is \textbf{bolded} and the second-highest is \underline{underlined}.}
\end{table*}

\begin{table*}[ht]
\centering
\begin{subtable}[t]{0.49\textwidth}
\centering
\scriptsize
\setlength{\tabcolsep}{2.5pt}
\begin{tabular}{lcccccccccccc}
\toprule
\makecell{Model and \\ Experiment} & U.1 & U.2 & U.3 & U.4 & U.5 & U.6 & U.7 & U.8 & U.9 & U.10 & Avg & Std \\
\midrule
\multicolumn{13}{l}{\textbf{gpt-4o-mini-2024-07-18}} \\
\quad \ZEROSHOT{} & 32 & 72 & 44 & \textbf{68} & \underline{64} & \underline{52} & \underline{68} & 52 & \textbf{56} & 56 & 56 & 12 \\
\quad \EPLABEL & \textbf{48} & 72 & \textbf{68} & 56 & 64 & 36 & 68 & 60 & \underline{48} & 56 & 58 & 11 \\
\quad \EPCRIT{} & \underline{44} & \underline{76} & \underline{68} & \underline{64} & 64 & \textbf{56} & 64 & \underline{72} & 44 & \underline{68} & 62 & 10 \\
\quad \SEMCRIT{} & 36 & 64 & 48 & 48 & 64 & 48 & 64 & 60 & 36 & 60 & 53 & 11 \\
\quad \EPSEMCRIT{} & 32 & \textbf{80} & 64 & 64 & \textbf{72} & 36 & \textbf{76} & \textbf{80} & 36 & \textbf{76} & 62 & 18 \\
\hline
\multicolumn{13}{l}{\textbf{gpt-5-2025-08-07}} \\
\quad \ZEROSHOT{} & 44 & \underline{76} & 64 & 36 & 64 & 44 & \textbf{72} & 52 & 52 & 52 & 56 & 12 \\
\quad \EPLABEL & \underline{68} & \textbf{84} & \textbf{96} & \underline{80} & 60 & \textbf{64} & \underline{64} & 88 & 52 & 64 & 72 & 13 \\
\quad \EPCRIT{} & \textbf{72} & 68 & \underline{80} & \textbf{92} & 64 & \underline{64} & 64 & \textbf{92} & \underline{60} & \textbf{76} & 73 & 11 \\
\quad \SEMCRIT{} & 56 & 48 & 68 & 60 & \underline{68} & 52 & 48 & 56 & \textbf{68} & 56 & 58 & 7 \\
\quad \EPSEMCRIT{} & 60 & 68 & 68 & 76 & \textbf{72} & 60 & 64 & \underline{92} & 52 & \underline{76} & 69 & 11 \\
\hline
\multicolumn{13}{l}{\textbf{gpt-oss-20b}} \\
\quad \ZEROSHOT{} & 24 & 60 & 52 & 52 & \textbf{68} & 44 & 52 & 64 & \underline{56} & 56 & 53 & 12 \\
\quad \EPLABEL & 32 & 68 & 52 & 52 & 44 & \textbf{48} & 52 & 72 & 48 & 68 & 54 & 12 \\
\quad \EPCRIT{} & \textbf{60} & 72 & \underline{72} & 48 & \underline{60} & \underline{48} & \underline{60} & \textbf{80} & 44 & 64 & 61 & 11 \\
\quad \SEMCRIT{} & \underline{52} & \textbf{76} & 52 & \underline{56} & 60 & 48 & \textbf{84} & 64 & 36 & \textbf{76} & 60 & 14 \\
\quad \EPSEMCRIT{} & 52 & \underline{76} & \textbf{76} & \textbf{60} & 48 & 44 & 44 & \underline{76} & \textbf{64} & \underline{72} & 61 & 13 \\
\hline
\multicolumn{13}{l}{\textbf{llama-4-scout}} \\
\quad \ZEROSHOT{} & \textbf{40} & 56 & 48 & \textbf{60} & 56 & 44 & 56 & 56 & \textbf{52} & 72 & 54 & 8 \\
\quad \EPLABEL & \underline{36} & \underline{72} & \underline{52} & 56 & \textbf{76} & \textbf{56} & 60 & \textbf{72} & 48 & \textbf{76} & 60 & 13 \\
\quad \EPCRIT{} & 36 & 64 & \textbf{64} & \underline{60} & \underline{64} & 44 & 60 & 64 & 48 & \underline{76} & 58 & 11 \\
\quad \SEMCRIT{} & 28 & \textbf{76} & 36 & 60 & 64 & \underline{48} & \textbf{64} & 68 & \underline{52} & 72 & 57 & 15 \\
\quad \EPSEMCRIT{} & 28 & 68 & 52 & 60 & 64 & 48 & \underline{64} & \underline{72} & 52 & 68 & 58 & 12 \\
\hline
\multicolumn{13}{l}{\textbf{qwen3-235b-a22b-instruct-2507}} \\
\quad \ZEROSHOT{} & 52 & 60 & 44 & 32 & 56 & 48 & \underline{64} & 60 & \textbf{64} & 48 & 53 & 10 \\
\quad \EPLABEL & \underline{64} & 64 & 76 & \underline{52} & 56 & \textbf{60} & 56 & \underline{84} & 60 & \underline{76} & 65 & 10 \\
\quad \EPCRIT{} & \textbf{72} & \underline{68} & \underline{84} & \textbf{64} & \textbf{64} & 44 & 56 & \textbf{92} & \underline{64} & \textbf{88} & 70 & 14 \\
\quad \SEMCRIT{} & 48 & 68 & 68 & 44 & \underline{60} & 52 & \textbf{68} & 72 & 64 & 72 & 62 & 10 \\
\quad \EPSEMCRIT{} & 64 & \textbf{72} & \textbf{88} & 52 & 60 & \underline{60} & 60 & 68 & 60 & 68 & 65 & 9 \\
\hline
\multicolumn{13}{l}{\textbf{qwen3-vl-235b-a22b-thinking}} \\
\quad \ZEROSHOT{} & 40 & \textbf{68} & 44 & 52 & \textbf{76} & 36 & \textbf{68} & 60 & \underline{56} & 56 & 56 & 12 \\
\quad \EPLABEL & \textbf{56} & \underline{68} & 64 & \textbf{80} & \underline{72} & 44 & \underline{68} & 80 & 52 & 64 & 65 & 11 \\
\quad \EPCRIT{} & \underline{56} & 64 & \textbf{72} & \underline{68} & 60 & \underline{48} & 52 & \textbf{92} & \textbf{60} & 64 & 64 & 12 \\
\quad \SEMCRIT{} & 56 & 64 & \underline{68} & 40 & 52 & 40 & 52 & 56 & 48 & \textbf{68} & 54 & 10 \\
\quad \EPSEMCRIT{} & 52 & 56 & 68 & 60 & 64 & \textbf{68} & 60 & \underline{84} & 56 & \underline{68} & 64 & 9 \\
\bottomrule
\end{tabular}
\caption{Per-user accuracy for Anime Preference.}
\label{tab:per_user_anime}
\end{subtable}
\begin{subtable}[t]{0.49\textwidth}
\centering
\scriptsize
\setlength{\tabcolsep}{2.5pt}
\begin{tabular}{lcccccccccccc}
\toprule
\makecell{Model and \\ Experiment} & U.1 & U.2 & U.3 & U.4 & U.5 & U.6 & U.7 & U.8 & U.9 & U.10 & Avg & Std \\
\midrule
\multicolumn{13}{l}{\textbf{gpt-4o-mini-2024-07-18}} \\
\quad \ZEROSHOT{} & 60 & 36 & 72 & \textbf{56} & 20 & \underline{68} & \underline{36} & 60 & \textbf{64} & 44 & 52 & 16 \\
\quad \EPLABEL & 36 & 52 & 92 & 24 & 24 & 20 & \textbf{48} & 64 & 48 & \textbf{48} & 46 & 21 \\
\quad \EPCRIT{} & 60 & \underline{64} & \underline{96} & \underline{52} & \textbf{40} & 68 & 32 & \textbf{72} & \underline{60} & 40 & 58 & 18 \\
\quad \SEMCRIT{} & \textbf{76} & 52 & 88 & 52 & 16 & \textbf{72} & 28 & 60 & 60 & \underline{48} & 55 & 20 \\
\quad \EPSEMCRIT{} & \underline{64} & \textbf{68} & \textbf{100} & 52 & \underline{40} & 64 & 36 & \underline{68} & 56 & 36 & 58 & 18 \\
\hline
\multicolumn{13}{l}{\textbf{gpt-5-2025-08-07}} \\
\quad \ZEROSHOT{} & 60 & 32 & 72 & 44 & 20 & 40 & \textbf{56} & \textbf{84} & 40 & 40 & 49 & 18 \\
\quad \EPLABEL & \textbf{64} & \underline{60} & 88 & \textbf{88} & \textbf{56} & \textbf{76} & \underline{52} & 72 & \underline{76} & 60 & 69 & 12 \\
\quad \EPCRIT{} & 56 & 56 & 88 & \underline{88} & 44 & \underline{72} & 44 & 72 & \textbf{88} & 48 & 66 & 17 \\
\quad \SEMCRIT{} & \underline{64} & 28 & \textbf{92} & 84 & 52 & 60 & 52 & \underline{80} & 44 & \textbf{80} & 64 & 19 \\
\quad \EPSEMCRIT{} & 60 & \textbf{64} & \underline{92} & 84 & \underline{56} & 72 & 32 & 72 & 72 & \underline{72} & 68 & 16 \\
\hline
\multicolumn{13}{l}{\textbf{gpt-oss-20b}} \\
\quad \ZEROSHOT{} & \underline{56} & \underline{48} & 76 & 36 & 28 & 60 & 44 & \textbf{76} & 56 & 40 & 52 & 15 \\
\quad \EPLABEL & 40 & 48 & \textbf{84} & 60 & 44 & 20 & 44 & \underline{76} & \underline{72} & \underline{52} & 54 & 18 \\
\quad \EPCRIT{} & 52 & \textbf{52} & 80 & \textbf{80} & \underline{52} & \textbf{72} & 48 & 72 & \textbf{80} & 52 & 64 & 13 \\
\quad \SEMCRIT{} & \textbf{64} & 32 & \underline{84} & 36 & 32 & 64 & \textbf{56} & 68 & 40 & 48 & 52 & 17 \\
\quad \EPSEMCRIT{} & 56 & 44 & 84 & \underline{72} & \textbf{60} & \underline{72} & \underline{52} & 68 & 68 & \textbf{60} & 64 & 11 \\
\hline
\multicolumn{13}{l}{\textbf{llama-4-scout}} \\
\quad \ZEROSHOT{} & 52 & \underline{40} & 68 & \textbf{48} & \textbf{32} & \textbf{76} & 40 & 60 & 44 & 36 & 50 & 14 \\
\quad \EPLABEL & \underline{56} & 40 & \textbf{96} & \underline{40} & 24 & 64 & 48 & 72 & \underline{68} & 40 & 55 & 20 \\
\quad \EPCRIT{} & \textbf{64} & 36 & 88 & 40 & \underline{32} & \underline{68} & \underline{52} & 72 & \textbf{76} & \textbf{48} & 58 & 18 \\
\quad \SEMCRIT{} & 44 & 32 & 92 & 36 & 20 & 64 & 48 & \underline{76} & 48 & 44 & 50 & 20 \\
\quad \EPSEMCRIT{} & 56 & \textbf{48} & \underline{96} & 36 & 20 & 64 & \textbf{56} & \textbf{80} & 60 & \underline{48} & 56 & 20 \\
\hline
\multicolumn{13}{l}{\textbf{qwen3-235b-a22b-instruct-2507}} \\
\quad \ZEROSHOT{} & 56 & 52 & 60 & 40 & 20 & \underline{64} & \underline{48} & 52 & \textbf{68} & \underline{56} & 52 & 13 \\
\quad \EPLABEL & 44 & 64 & 84 & 40 & 48 & 36 & 36 & 56 & 56 & 56 & 52 & 14 \\
\quad \EPCRIT{} & 56 & 48 & \underline{92} & \underline{64} & 60 & \textbf{68} & 24 & 64 & 60 & 48 & 58 & 16 \\
\quad \SEMCRIT{} & \textbf{60} & \textbf{72} & 92 & 56 & \underline{72} & 56 & \textbf{52} & \underline{68} & 40 & \textbf{60} & 63 & 13 \\
\quad \EPSEMCRIT{} & \underline{60} & \underline{72} & \textbf{96} & \textbf{68} & \textbf{76} & 56 & 32 & \textbf{72} & \underline{64} & 44 & 64 & 17 \\
\hline
\multicolumn{13}{l}{\textbf{qwen3-vl-235b-a22b-thinking}} \\
\quad \ZEROSHOT{} & 52 & 44 & 80 & 40 & 20 & 64 & 44 & 60 & 52 & 36 & 49 & 16 \\
\quad \EPLABEL & \underline{56} & \textbf{64} & 88 & \textbf{84} & \textbf{64} & 64 & 40 & 60 & 60 & 36 & 62 & 15 \\
\quad \EPCRIT{} & \textbf{60} & \underline{60} & 88 & \underline{84} & \underline{60} & \underline{72} & \textbf{48} & \textbf{80} & \textbf{76} & 44 & 67 & 14 \\
\quad \SEMCRIT{} & 52 & 48 & \underline{92} & 80 & 48 & \textbf{84} & 40 & 56 & 68 & \textbf{52} & 62 & 17 \\
\quad \EPSEMCRIT{} & 56 & 48 & \textbf{96} & 80 & 48 & 68 & \underline{48} & \underline{72} & \underline{76} & \underline{48} & 64 & 16 \\
\bottomrule
\end{tabular}
\caption{Per-user accuracy for Movie Preference.}
\label{tab:per_user_movie}
\end{subtable}
\caption{For each model and user, the highest score is \textbf{bolded} and the second-highest is \underline{underlined}.}
\end{table*}

%% file: tex/artifact_appendix.tex
\section{Artifact Replication}
The provided code hosted at \url{https://github.com/megagonlabs/critique-learning} reproduces every quantitative result in the paper, including all tables and numerical claims. It ships (i)~the full pipeline library
(performance agent, critic agent, FAISS + sentence-transformer RAG, all
prompt templates), (ii)~the exact train/test splits used in the paper,
(iii)~the full output logs of every experiment, and (iv)~a single driver
that reproduces all results end-to-end plus one-command analysis scripts
that print every table. The analysis path is self-contained and requires
no API keys; the full re-run calls hosted LLMs via OpenAI and Fireworks.ai.

We are applying for all three badges. The artifact is publicly archived (Available), ships a documented end-to-end pipeline with a minimal working example (Functional), and reproduces every quantitative claim and table in the paper from shipped logs or a full re-run (Reproduced).

Recommended path for reviewers: run E0 (30s, no API access) to verify all claims against shipped logs; optionally run E1 (<10 min, \$0.20) to verify the pipeline is functional. E2 is the full reproduction path but is impractical at ~400h / ~\$600.

\subsection{Description \& Requirements}

\subsubsection{How to access}
The artifact will be archived at Zenodo by the end of artifact evaluation and is currently mirrored at
\url{https://github.com/megagonlabs/critique-learning}.
Licensed under BSD~3-Clause. Dataset splits under \texttt{data\_samples/}
are transformations of upstream datasets and inherit their upstream
licenses; see the repo-level \texttt{README.md}
for per-dataset attribution.

\subsubsection{Hardware dependencies}
No GPU required. All LLM inference is routed to hosted APIs. The only
local model is the sentence-transformer embedder used for RAG
(\texttt{blevlabs/stella\_en\_v5}, $\approx$1.5B~params); CPU execution
works, and \texttt{torch} picks up a GPU automatically if one is present
but there is no GPU-only code path.

\subsubsection{Software dependencies}
Ubuntu~22.04.2, Python~3.10.16, and the pinned package set in
\texttt{requirements.txt}. No containers, privileged
operations, or kernel modules.

\subsubsection{Benchmarks / datasets}
Seven datasets ship pre-split under data\_\allowbreak samples/:
multi\_\allowbreak condition\_\allowbreak ranking\_\allowbreak multichoice,
nfcorpus\_\allowbreak short\_\allowbreak questions, pubmed, and the four
preference tasks \{anime, book, movie, steam\}\_\allowbreak sample\_\allowbreak \{1..10\} (ten independent users each,
averaged in the paper). No downloading required for reproduction.

\subsubsection{API access (re-run experiments only)}
{\small
\begin{itemize}
  \item \texttt{OPENAI\_API\_KEY} --- for \texttt{gpt-4o-mini-2024-07-18} and  
  
        \texttt{gpt-5-2025-08-07}.
  \item \texttt{FIREWORKS\_API\_KEY} (+\texttt{FIREWORKS\_BASE\_URL}) ---
        for \texttt{gpt-oss-20b}, \texttt{llama-4-scout},
        \texttt{qwen3-235b-a22b-instruct-2507},
        \texttt{qwen3-vl-\allowbreak 235b-\allowbreak a22b-thinking}.
\end{itemize}
}
The analysis-only path E0 requires no API access.

\subsection{Installation}
{\small
\begin{verbatim}
git clone https://github.com/megagonlabs/critique-learning.git 
cd critique-learning
python3.10 -m venv .venv
source .venv/bin/activate
pip install -r requirements.txt
\end{verbatim}
}
First invocation downloads and caches the sentence-transformer embedder
via \texttt{huggingface\_hub}.

\subsection{Major claims}
\label{sec:claims}
\begin{description}
\item[C1 (accuracy):] The \texttt{EP+SEM\_CRIT} strategy
yields an average $+8.1$\,pp over zero-shot and $+4.6$\,pp over
label-only RAG (\texttt{EP\_LABEL}) across all six models and seven datasets.
\item[C2 (gain significance):] EP+SEM\_CRIT improves over EP\_LABEL on 32 of 42 (model, dataset) points (sign-test p < 0.001), and over zero\_shot on 38 of 42 points (sign-test p $<$ $10^{-7}$).
\item[C3 (reasoning-token reduction):] On reasoning models, critique-based memory cuts thinking tokens by an average of 31.95\%.
\item[C4 (suggestibility):] Suggestibility scores correlate with EP+SEM\_CRIT gain over zero\_shot (Spearman coefficient=0.402, p=0.008).
\end{description}

All of the above numbers are present verbatim under the \texttt{Summary statistics} section of the report generated by:
\begin{verbatim}
python scripts/reproduce_paper_tables.py
\end{verbatim}

\subsection{Experiments}

\noindent\textbf{E0 (Analysis only; no API access; $\approx$30\,s):}
\begin{verbatim}
python scripts/reproduce_paper_tables.py
\end{verbatim}
prints every table cited in the paper against the shipped log folders: main\_results, train\_size\_ablation, and reflection\_parts\_ablation.
This alone verifies C1--C4 from the provided logs.

\noindent\textbf{E1 (Basic test, $<$10\,min, $\approx$\$0.20)}
\begin{verbatim}
export OPENAI_API_KEY=sk-...
python scripts/run_pipeline.py \
  --dataset data_samples/anime_sample_1 \
  --model  gpt-4o-mini-2024-07-18
python scripts/analyze_logs.py logs/<timestamp>
\end{verbatim}
This runs all five strategies and the consistency probe on one user and
prints an accuracy table, confirming the pipeline is functional.

\noindent\textbf{E2 (Full reproduction from scratch; $\approx$400\,h
wall-clock, $\approx$\$600 API spend):}
\begin{verbatim}
python scripts/run_all_experiments.py
python scripts/reproduce_paper_tables.py \
  --logs logs/<wallclock>
\end{verbatim}
The driver runs all four stages (main pipeline, vary-$K$, train-size
ablation, reflection-parts ablation) of experiments, outputs results into one timestamped directory, and displays results in formats compatible with paper's claims.

\subsection{Comparing to the paper}
Every analyzer prints numbers in the same units, shape, and ordering as
the corresponding paper table, with row/column headers matching the
paper's model and dataset names. 

\begin{verbatim}python scripts/reproduce_paper_tables.py\end{verbatim}
Upon running the above command, all paper claims and tables are recreated from existing logs, and match the numbers presented in the paper exactly.

If re-running experiments, small deviations on rerun experiments from numbers in the paper are expected due to LLM non-determinism
even at temperature~0, particularly on the 25-question preference
splits. However, aggregate claims (C1--C4) are robust.

\subsection{Expected anomalies}
Reasoning-model outputs contain \texttt{ Error: token limit exceeded}
rows; these are counted as wrong by design (see \S "Unusual behavior" in
\texttt{README.md}) and their reduction is itself part of claim C3.

Parse failures return a random wrong answer by design to prevent
malformed JSON responses from silently inflating accuracy.